\documentclass[10pt,twocolumn,letterpaper]{article}

\usepackage{cvpr}              

\usepackage{multirow}
\usepackage{colortbl}
\usepackage{graphicx}
\usepackage{booktabs}
\usepackage{xcolor}
\usepackage{array}
\usepackage{makecell}
\usepackage{amssymb}
\usepackage{pifont}

\newcommand{\cmark}{\ding{51}}%
\newcommand{\xmark}{\ding{55}}%

\newcommand{\tabincell}[2]{\begin{tabular}{@{}#1@{}}#2\end{tabular}}

\newcommand{\myred}[1]{\textcolor{red}{{#1}}}
\newcommand{\myblue}[1]{\textcolor{blue}{{#1}}}
\definecolor{darkgreen}{RGB}{0,150,0}
\newcommand{\mygreen}[1]{\textcolor{darkgreen}{{#1}}}
\definecolor{cvprblue}{rgb}{0.21,0.49,0.74}
\usepackage[pagebackref,breaklinks,colorlinks,citecolor=cvprblue]{hyperref}
\newcommand{\gray}[1]{\textcolor{gray}{#1}}

\newcommand{\symbolfootnote}[1]{%
  \renewcommand{\thefootnote}{$\dagger$}
  \footnotetext{#1}
  \renewcommand{\thefootnote}{\arabic{footnote}}
}

\def\paperID{6416} 
\def\confName{CVPR}
\def\confYear{2025}

\title{\raisebox{-0.1em}{\includegraphics[height=0.9em]{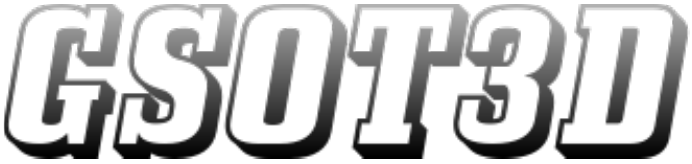}}: Towards Generic 3D Single Object Tracking in the Wild}

\author{Yifan Jiao$^{1,2}$\; Yunhao Li$^{1,2}$ \; Junhua Ding$^{3}$ \; Qing Yang$^{3}$ \; Song Fu$^{3}$ \; Heng Fan$^{3\dagger}$ \; Libo Zhang$^{1\dagger}$\\
$^{1}$University of Chinese Academy of Sciences \\
$^{2}$Institute of Software Chinese Academy of Sciences \;\;\;
$^{3}$University of North Texas\\
}

\begin{document}

\twocolumn[{%
\renewcommand\twocolumn[1][]{#1}%
\maketitle
\begin{center}
    \centering
    \captionsetup{type=figure}
    \includegraphics[width=0.975\textwidth]{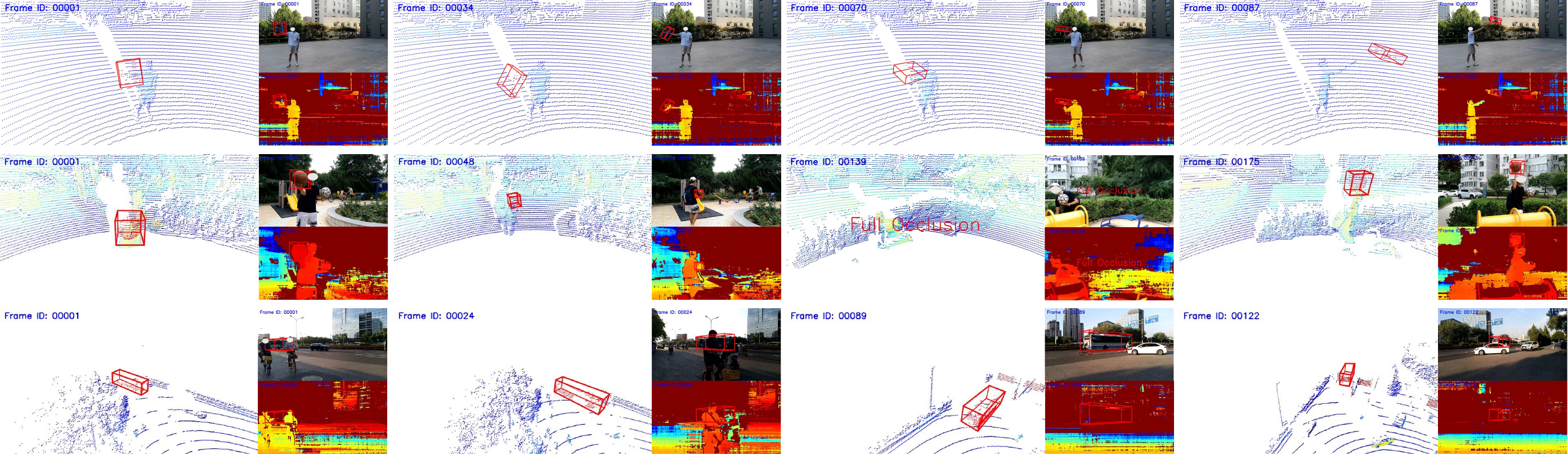}   
    \captionof{figure}{Demonstration of a few sequence samples from our GSOT3D. Each sequence is offered with multiple modalities, including \emph{point cloud}, \emph{RGB image}, and \emph{depth}, supporting different 3D SOT tasks. \emph{Best viewed in color and by zooming in for all figures in the paper}.} 
    \label{fig:exa}
\end{center}%
}]

\maketitle

\begin{abstract}
In this paper, we present a novel benchmark, \textbf{GSOT3D}, that aims at facilitating development of generic 3D single object tracking (SOT) in the wild. Specifically, GSOT3D offers 620 sequences with 123K frames, and covers a wide selection of 54 object categories.\symbolfootnote{Equal advising and co-last authors.} Each sequence is offered with multiple modalities, including the point cloud (PC), RGB image, and depth. This allows GSOT3D to support various 3D tracking tasks, such as single-modal 3D SOT on PC and multi-modal 3D SOT on RGB-PC or RGB-D, and thus greatly broadens research directions for 3D object tracking. To provide high-quality per-frame 3D annotations, all sequences are labeled manually with multiple rounds of meticulous inspection and refinement. To our best knowledge, GSOT3D is the largest benchmark dedicated to various generic 3D object tracking tasks. To understand how existing 3D trackers perform and to provide comparisons for future research on GSOT3D, we assess eight representative point cloud-based tracking models. Our evaluation results exhibit that these models heavily degrade on GSOT3D, and more efforts are required for robust and generic 3D object tracking. Besides, to encourage future research, we present a simple yet effective generic 3D tracker, named PROT3D, that localizes the target object via a progressive spatial-temporal network and outperforms all current solutions by a large margin. By releasing GSOT3D, we expect to advance further 3D tracking in future research and applications. Our benchmark and model as well as the evaluation results will be publicly released at our webpage \href{https://github.com/ailovejinx/GSOT3D}{https://github.com/ailovejinx/GSOT3D}.
\end{abstract}

\section{Introduction}
\label{sec:intro}
As one of the most crucial problems in 3D computer vision, 3D single object tracking (SOT) aims to localize the desired target with a sequence of 3D bounding boxes, given its state in the first frame. Due to its key roles in many applications, such as intelligent vehicles, mobile robotics, navigation, etc, 3D object tracking has gained extensive attention in the past decade with many models proposed (\eg,~\cite{asvadi20163d,bibi20163d,giancola2019leveraging,qi2020p2b,zheng2021box}).

Current research mainly focuses on the point cloud (PC)-based 3D tracking. Relying on popular autonomous driving benchmarks (\eg, KITTI~\cite{geiger2012we} and NuScenes~\cite{caesar2020nuscenes}), numerous deep 3D trackers have been proposed and demonstrated state-of-the-art results (\eg,~\cite{xu2023cxtrack,xu2023mbptrack,nie2024towards,wu20253d}). Despite such progress, further development of \emph{generic} 3D SOT is heavily \emph{restricted} by currently adopted benchmarks due to several reasons: \textbf{(1)} \emph{\textbf{limited object classes}}. To achieve general tracking capacity, a 3D tracker is expected to learn with sequences from a large set of categories during training. However, existing datasets for 3D SOT (\eg,~\cite{geiger2012we,caesar2020nuscenes}), specially designed for autonomous driving, comprise \emph{very few} available categories (\eg, 8 in~\cite{geiger2012we} and 23 in~\cite{caesar2020nuscenes}) for tracking, making them \emph{inadequate} for designing generic 3D trackers. \textbf{(2)} \emph{\textbf{constrained scenarios}}. In applications, a general tracker should be able to localize the target object under various scenarios, which requires it to be trained and assessed with sequences collected from diverse environments. Yet current datasets, due to their own specific aims, only offer sequences from the traffic scenario and thus are \emph{unsuitable} for general tracking. \textbf{(3)} \textbf{\emph{restricted degrees of freedom (DoF)}}. For generic 3D tracking, a tracker needs to handle objects with arbitrary pose and size, often described with 9DoF consisting of 6D pose and 3D size. Nonetheless, currently used datasets~\cite{geiger2012we,caesar2020nuscenes} comprise only targets of 7DoF, including 4D pose and 3D size, and thus are \emph{undesirable} for developing general trackers locating arbitrary-pose objects.

It is worth noting that, besides the PC-based 3D SOT, the above autonomous driving datasets (\eg,~\cite{geiger2012we,caesar2020nuscenes}) can also be used for developing multi-modal, \ie, RGB-PC, tracking by integrating point clouds and RGB images. Nevertheless, the aforementioned issues still exist, and therefore, limit the further development of generic 3D object tracking.

In addition to PC-based single- or multi-modal solutions, another direction that is more affordable is to leverage RGB and depth information for 3D tracking. For such a goal, a recent dataset~\cite{yang2022towards} has been introduced by collecting RGB-D sequences from diverse categories and annotating each one with 9DoF 3D boxes. However, it is \emph{limited} by its relatively \textbf{\emph{small scale}}. In order to effectively train and reliably assess deep 3D trackers, it is desirable to have plenty of sequences in a dataset. Nonetheless in~\cite{yang2022towards}, there is a total of only 300 sequences with 36K frames, which might be \emph{insufficient} for large-scale learning and evaluation of deep 3D trackers. 

\vspace{0.7em}
\noindent
\textbf{Contributions.} To alleviate limitations in existing 3D SOT benchmarks and offer a versatile platform for 3D tracking, we introduce a high-quality benchmark, \emph{\textbf{GSOT3D}}, which is dedicated to diverse generic 3D object tracking.

Specifically, our GSOT3D consists of 620 sequences and provides more than 123K frames in total. In order to ensure the diversity of GSOT3D, these sequences are carefully collected from a wide selection of 54 object classes from various environments. For each sequence in GSOT3D, multiple modalities, including the \emph{point cloud (PC)}, \emph{RGB image}, and \emph{depth}, are offered using different sensors (see examples in Fig.~\ref{fig:exa}). This allows GSOT3D to support different 3D tracking tasks, comprising the \emph{single-modal} 3D SOT on PC and \emph{multi-modal} 3D SOT on  RGB-PC or RGB-D, and therefore broadens the research directions in 3D tracking. For precise dense annotations, all the sequences in GSOT3D are manually labeled using 9DoF 3D bounding boxes with multiple rounds of inspection and refinement. To our best knowledge, GSOT3D is, to date, the \emph{largest} benchmark dedicated to generic 3D object tracking. Besides, it is the \emph{first}  benchmark, to date, that simultaneously supports different single- and multi-modal 3D SOT tasks.  

Compared with existing benchmarks (\eg,~\cite{geiger2012we,caesar2020nuscenes}) with a few object classes for 3D SOT on PC and RGB-PC in traffic scene, GSOT3D is more \emph{diverse} by containing 54 categories and various scenarios, making it more favorable for generic 3D tracking. Moreover, compared to~\cite{yang2022towards} consisting of 300 sequences with 36K frames for RGB-D 3D tracking, GSOT3D is \emph{larger} by providing 620 sequences (2$\times$ larger) with 123K frames (3$\times$ larger), and hence more desirable for large-scale learning and evaluation of deep 3D tracking.

In order to understand how existing 3D trackers perform and to provide comparisons for future research, we assess 8 representative PC-based tracking methods. Please note that, compared to 2D generic object tracking, there are \emph{not} many open-sourced 3D trackers and most methods are PC-based. For this reason, we finally include 8 PC-based trackers, that are representative and provide executable implementations, for evaluation. Our evaluation reveals that, not surprisingly, all current models degrade severely on the more challenging GSOT3D, which demonstrates the difficulty in achieving generic 3D tracking in the real-world, and more efforts are needed for future improvements. 

Moreover, to facilitate research on GSOT3D, we present a simple but effective generic 3D tracker, dubbed \textbf{\emph{PROT3D}}, for \emph{class-agnostic} 3D tracking on point clouds. The core of PROT3D is a progressive spatial-temporal architecture containing multiple stages. In each stage, target localization is performed by spatial-temporal matching with Transformer, and the result is applied to refine search region feature. The refined search region feature from one stage is forwarded to next stage for further improvements, and tracking result is generated after the final stage. This way, PROT3D gradually learns more discriminative features via progressive feature refinement, making it capable of handling more complex scenarios for generic tracking. It is worth noticing, unlike current trackers predicting a 7DoF box, our PROT3D produces a 9DoF box for more precise tracking. Despite its simplicity, PROT3D outperforms all other methods, and expects to provide a reference for future research.

In summary, our contributions are as follows: \ding{171} We propose a new benchmark GSOT3D comprising 620 sequences with more than 123K frames to facilitate 3D object tracking; \ding{170} GSOT3D provides multiple modalities to each sequence, making it a versatile platform for various research directions in 3D tracking;  \ding{168} We evaluate eight representative trackers to understand their performance and to offer comparisons to future research; \ding{169} We present a simple yet effective tracker, PROT3D, to encourage future research on GSOT3D.

\renewcommand{\arraystretch}{1.05}
\begin{table*}[!t]\small
  \centering
  \caption{Detailed comparison of our GSOT3D with existing 3D SOT benchmarks. O: Outdoor, I: Indoor, PC: Point cloud, D: Depth. Please notice that, we gray KITTI and NuScenes, as they are \emph{not} specifically developed for 3D single object tracking. $\P$: Based on the information provided in the original paper~\cite{yang2022towards}, there are 44 object categories in total in Track-it-in-3D.}\vspace{-2mm}
  \resizebox{0.99\textwidth}{!}{
    \begin{tabular}{rcccccccccccc}
    \Xhline{1.2pt}
    \multirow{2}[0]{*}{Benchmark}  & \multirow{2}[0]{*}{Where} & \multirow{2}[0]{*}{\tabincell{c}{Total\\Sequences}} & \multirow{2}[0]{*}{\tabincell{c}{Total\\Frames}} & \multirow{2}[0]{*}{\tabincell{c}{Avg.\\Length}} & \multirow{2}[0]{*}{\tabincell{c}{Object\\Classes}} & \multirow{2}[0]{*}{\tabincell{c}{Data\\Scenarios}} & \multicolumn{3}{c}{Modality} & \multicolumn{3}{c}{3D SOT Task on} \\
    \cmidrule(lr){8-10} \cmidrule(lr){11-13}
    &       &     &   &  &          &       & RGB    & PC   & Depth & PC & RGB-PC & RGB-D \\
    \hline\hline
    \gray{\textbf{KITTI}~\cite{geiger2012we}}  & \gray{CVPR'2012}  & \gray{21}    & \gray{15K}  & - & \gray{8}     & \gray{O}     &  \gray{\cmark}     &  \gray{\cmark}     &  \gray{\xmark}     &  \gray{\cmark}     &  \gray{\cmark}     & \gray{\xmark} \\
    \gray{\textbf{NuScenes}~\cite{caesar2020nuscenes}}  & \gray{CVPR'2020}  & \gray{1,000}  & \gray{40K} & - & \gray{23}     & \gray{O}     & \gray{\cmark}     &  \gray{\cmark}     &  \gray{\xmark}     &  \gray{\cmark}     &  \gray{\cmark}     & \gray{\xmark}  \\
    \textbf{Track-it-in-3D}~\cite{yang2022towards}  & ECCV'2022  & 300   & 36K & 120 & 44$^{\P}$   & I \& O  &  \cmark     &  \xmark     &  \cmark     &  \xmark     &  \xmark     & \cmark  \\
    \hline
    \rowcolor[HTML]{dee9fc} \textbf{GSOT3D} (ours) & -   & 620   & 123K & 198 & 54    & I \& O  & \cmark     &  \cmark     &  \cmark     &  \cmark     &  \cmark     & \cmark  \\
    \Xhline{1.2pt}
    \end{tabular}%
  \label{tab:benchmark}%
  }
\vspace{-3mm}
\end{table*}%

\section{Related Work}
\label{sec:related}

\textbf{Benchmarks for 3D Single Object Tracking.} Datasets are crucial for 3D single object tracking by providing platforms for training and assessment. Currently, the popular datasets, particularly for 3D tracking on point cloud, are mainly borrowed from the autonomous driving benchmarks, including  KITTI~\cite{geiger2012we} and NuScenes~\cite{caesar2020nuscenes}. Specifically, KITTI comprises 21 sequences with 15K frames, and each one is offered with point clouds and RGB images. Similar to KITTI but with a larger size, NuScenes comprises 1,000 sequences with 40K frames. Since KITTI and NuScenes are originally designed for autonomous driving, they usually need appropriate conversions before being used for 3D SOT. Besides KITTI and NuScenes for point cloud-related 3D SOT, the work of~\cite{yang2022towards} recently proposes a new benchmark, named Track-it-in-3D, dedicated to RGB-D-based 3D object tracking. It contains 300 sequences with 36K frames, collected from 44 classes. Each sequence is annotated with 9DoF 3D boxes for more precise generic 3D object tracking.

Despite the above benchmarks, the further development of 3D SOT remains constrained by the limitations discussed earlier, which motivates our GSOT3D in this work, a versatile dataset dedicated to different generic 3D tracking tasks. Tab.~\ref{tab:benchmark} compares our GSOT3D with other datasets in detail.

\vspace{0.3em}
\noindent
\textbf{3D Object Tracking Algorithms.} 3D tracking has received extensive attention in the past decade. Most recent research focuses on point cloud-based 3D object tracking. The seminal work of~\cite{giancola2019leveraging} adopts a Siamese network that explores the shape completion for 3D tracking on point clouds. In order to improve the efficiency and enhance the performance, the work of~\cite{qi2020p2b} introduces an end-to-end framework that integrates target proposal and verification for 3D tracking. The method of~\cite{zheng2021box} leverages prior information from the target box to enhance features for improvement. The work of~\cite{zheng2022beyond} explores the motion cues from a sequence for 3D tracking, displaying promising results. The method of~\cite{hui20213d} proposes to improve tracking performance on sparse point clouds by learning shape-aware features and localizing the target from the dense bird’s eye view (BEV) feature maps, boosting the tracking results. More recently, inspired by~\cite{vaswani2017attention}, the Transformer has been extensively used for 3D tracking, showing excellent results~\cite{shan2021ptt,zhou2022pttr,guo2022cmt,hui20223d,xu2023cxtrack,ma2023synchronize,xu2023mbptrack,wu20253d,nie2024towards,wuboosting}. 

Besides 3D tracking on point clouds, another direction is to leverage RGB and depth information for 3D SOT. The work of~\cite{bibi20163d} introduces a part-based 3D tracker using sparse learning. In~\cite{yang2022towards}, a Siamese network is proposed to fuse the RGB and depth information for RGB-D 3D tracking.

\vspace{0.3em}
\noindent
\textbf{Generic 2D Tracking Datasets.} Our GSOT3D in this work is inspired, to some extent, by existing generic 2D tracking datasets. Early datasets, such as~\cite{wu2013online,liang2015encoding,li2015nus,kristan2016novel,galoogahi2017need,muller2018trackingnet}, mainly aim at evaluating and comparing the tracking performance, and are usually small-scale. Later, to facilitate development of generic tracking in deep learning era, several large-scale tracking datasets (\eg,~\cite{fan2019lasot,huang2019got,wang2021towards,peng2024vasttrack,muller2018trackingnet}) have been developed by offering abundant videos. Particularly, these large benchmarks often include a diverse selection of categories, well enhancing the generalization ability of deep trackers.

Sharing a similar goal with current large-scale 2D tracking benchmarks, GSOT3D aims at providing sufficient sequences from rich classes for generic 3D tracking. It is worthy to note that, compared to current large-scale 2D tracking benchmarks (\eg,~\cite{fan2019lasot,huang2019got,wang2021towards,peng2024vasttrack,muller2018trackingnet}) with over a thousand or tens of thousands videos, GSOT3D is relatively smaller due to the extreme difficulty in collecting sequences and annotating them using the 9DoF bounding boxes. That being said, GSOT3D to date is still the largest dataset that is dedicated to generic 3D single object tracking.

\begin{figure*}[!t]
\centering
\includegraphics[width=\linewidth]{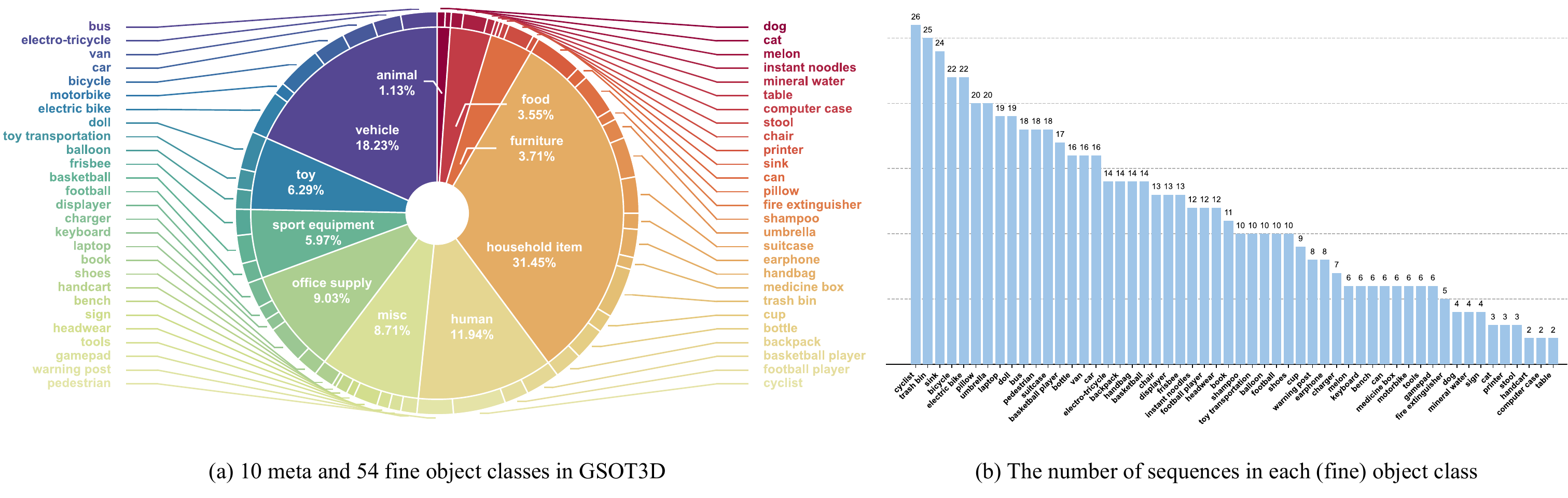}
\caption{Illustration of category organization in GSOT3D (image (a)) and its distribution of sequence number in each classes (image (b)).}
\label{fig:class_statistic}
\vspace{-2mm}
\end{figure*}

\section{The Proposed GSOT3D Benchmark}

\subsection{Construction Principle}

GSOT3D aims at serving as a \emph{versatile} platform to facilitate different 3D tracking tasks with sufficient sequences and rich classes as well as high-quality annotations. To this end, we follow several principles when constructing GSOT3D:

\begin{itemize}
\item \emph{Rich Object Class.} To achieve generic tracking, it is desirable to encompass diverse object categories in both training and evaluation. For this purpose, the new benchmark is expected to cover at least 50 categories, including common targets suitable for 3D tracking in our daily life.
\item \emph{Different 3D Tracking Tasks.} To broaden research directions in 3D SOT, multiple modalities should be provided for the sequences, allowing researchers to flexibly explore various 3D tracking tasks using different input types (single or multiple modalities) based on their specific needs. 
\item \emph{Appropriate Scale.} To effectively train and evaluate deep trackers, sufficient sequences are needed for a benchmark. Considering the difficulty in collecting and labeling data for 3D tracking, we hope to gather at least 600 sequences with over 100K frames in the new benchmark. 
\item \emph{Precise Annotation.} Precise annotation is important for a dataset. Thus, we manually label every frame in GSOT3D using more precise 9DoF 3D boxes, and carefully inspect and refine the annotations to ensure high quality.
\end{itemize}


\subsection{Data Acquisition.} 

\textbf{Data Acquisition Platform.} To collect data for GSOT3D, we build a mobile robotic platform based on the popular Clearpath Husky A200, and equip it with multiple sensors, including a 64-beam LiDAR, a depth camera, and an RGB camera. All these sensors have been calibrated and synchronized, and the system allows for stably outputting point clouds and (RGB and depth) images synchronized at 10 or 20 frames per second (\emph{fps}). In this work, we choose 20 fps, because this provides more dense temporal information. For more details and a picture of our platform, please kindly refer to our \textbf{supplementary material} due to space limitation.

\vspace{0.3em}
\noindent
\textbf{Collection of Sequences.} Different from current 2D tracking datasets that source videos from Internet, we record sequences using our mobile robot from diverse natural scenarios such as street, park, office, house, hall, etc. To start with, we first determine meta classes of GSOT3D that are suitable for 3D tracking. Please note, some classes that are common in 2D tracking, such as fish and bird, are \emph{not suitable} for 3D tracking due to difficulty in data collection and annotation. In GSOT3D, we select 10 meta classes, including \emph{furniture}, \emph{human}, \emph{vehicle}, \emph{household item}, \emph{office supply}, \emph{food}, \emph{animal}, \emph{sport equipment}, \emph{toy}, and \emph{misc}. Under each meta category, we further choose 54 fine classes. Fig.~\ref{fig:class_statistic} (a) shows 10 meta and 54 fine categories in GSOT3D, and (b) the distribution of the number of sequences in each fine category.

After determining the categories, we use our mobile platform to record sequences. To ensure the recorded sequences are suitable for 3D tracking, we invite several experts (students who work on 2D and 3D tracking) for data collection. Afterwards, each sequence is inspected by the expert group and inappropriate parts or intuitable sequences are removed. Finally, we compile a new benchmark which is dedicated to 3D SOT by comprising 620 multi-modal (\ie, RGB image, point cloud, and depth) sequences with over 123K frames from 54 object classes. The average sequence length of our GSOT3D is 198. Compared to the recent dataset~\cite{yang2022towards} containing 300 sequences for RGB-D 3D SOT, GSOT3D is 2$\times$ larger in size by including 620 sequences. A detailed comparison of GSOT3D with other datasets is in Tab.~\ref{tab:benchmark}.

\subsection{Annotation}

To ensure high quality of annotations in GSOT3D, we manually label each frame. Specifically, for each frame, we annotate the target with the tightest 9DoF 3D box to cover its any visible part if it shows up; otherwise an absence label, either \emph{full occlusion} or \emph{out-of-view}, is assigned to the frame. similar to the strategy as in 2D tracking datasets~\cite{fan2019lasot,fan2021lasot}. 

With the above strategy, we compile an annotation team, composed of several experts and a qualified labeling group, and use a multi-step mechanism for annotation. In the first step, the experts label the initial target in each sequence, and volunteers start to work on annotating the sequences. Then, in the second step, the experts work to verify the complected annotations in the first step. If the annotation is not unanimously agreed by the experts, it is sent back to the original annotator for refinement in the third step. During the whole annotation process, the verification and refinement from the second and third steps are repeated for multiple rounds until all annotations pass the verification, which ensures the high quality of our annotations. Fig.~\ref{fig:exa} displays several examples of our annotation in GSOT3D. Due to the limited space, we include the details about annotation tool, reliability analysis, and more statistics in the \textbf{supplementary material}. 

\subsection{Attributes}

In order to enable in-depth analysis, we annotate sequences in GSOT3D with 7 attributes, comprising \emph{invisibility} (INV), which is assigned when the target is partially or fully invisible due to occlusion and/or out of view, \emph{deformation} (DEF), which is assigned when the target is deformable, \emph{fast motion} (FM), which is assigned when target moves larger than half size of its bounding box, \emph{rotation} (ROT), which is assigned when target rotates in the view, \emph{scale variation} (SV), which is assigned when the ratio of the 3D box is beyond [0.75, 1.5], \emph{Similar Distractors} (SD), whish is assigned when there exist similar targets in the view, and \emph{Sparsity} (SPA), which is assigned when target information (point cloud or appearance) is sparse, \ie, the target region contains less than 50 points on PC or 1,000 pixels on RGB or depth. For each sequence, a 7D binary vector is used to indicate the presence of an attribute: ``1'' for presence, and ``0'' otherwise. 

\begin{figure}[!t]
\centering
\includegraphics[width=0.9\linewidth]{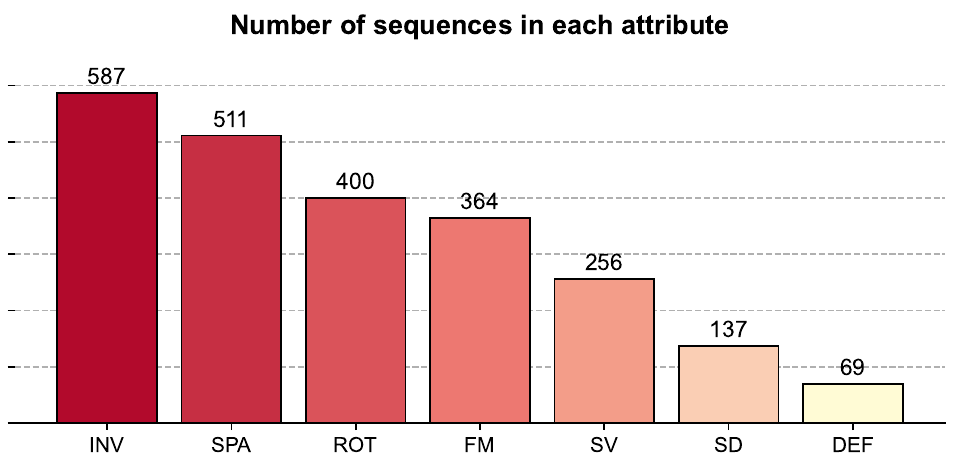}
\caption{Distribution of videos per attribute.}
\label{fig:attribute}
\vspace{-1.5mm}
\end{figure}

Fig.~\ref{fig:attribute} demonstrates the distribution of attributes. We can see that the most common attribute is INV, which may cause severe feature degradation for tracking. Besides, SPA and ROT frequently happen in sequences. We also notice, there are a few sequences involved with DEF, as some targets belonging to the human and animal meta classes are non-rigid, making the localization of them more challenging.

\subsection{Dataset Split, Evaluation Protocol, and Tasks}

\begin{table}[!t]\small
  \centering
  \caption{ Comparison of training and test sets of GSOT3D.}\vspace{-2mm}
    \begin{tabular}{ccccc}
    \Xhline{1.2pt}
          & \tabincell{c}{Total\\Sequences} & \tabincell{c}{Total\\Frames} & \tabincell{c}{Ave.\\Frames} & \tabincell{c}{Object\\Classes} \\
    \hline\hline
    GSOT3D$_{\text{Tra}}$ & 435   & 83,950   & 193   & 54 \\
    GSOT3D$_{\text{Tst}}$  & 185   & 39,740   & 215   & 54 \\
    \Xhline{1.2pt}
    \end{tabular}%
  \label{tab:split}%
  \vspace{-3mm}
\end{table}%

\textbf{Dataset Split.} Our GSOT3D includes 620 multi-modal sequences, and we adopt the 70/30 principle to generate training and test splits. In specific, 435 sequences are utilized in the training set named  GSOT3D$_{\text{Tra}}$, and the rest 185 for test set dubbed GSOT3D$_{\text{Tst}}$. Both GSOT3D$_{\text{Tra}}$ and GSOT3D$_{\text{Tra}}$ contain all the 54 object categories. In the dataset split, we try our best to make the distributions of these two sets close to each other. Tab.~\ref{tab:split} displays the comparison of GSOT3D$_{\text{Tra}}$ and GSOT3D$_{\text{Tst}}$, and the detailed splits will be released on our project paper together with our data and other materials. 

\vspace{0.45em}
\noindent
\textbf{Evaluation Protocol.} Inspired by~\cite{huang2019got}, we leverage mean Average Overlap (\textbf{mAO}) and mean Success Rate (\textbf{mSR}) for evaluation. mAO is computed by averaging the class-wise overlaps, \ie, 3D Intersection over Union (or 3D IoU), between all tracking results and the groundtruth, while mSR measures class-wise percent of successful frames in which 3D IoU is larger than a threshold (\eg, 0.5 or 0.75). The details of how to compute mAO and mSR as well as 3D IoU for different cases (non-symmetric and symmetric objects) can been seen in the \textbf{supplementary material}.

Please notice here, we do \emph{not} utilize the precision metric as in previous studies for evaluation, because the precision, that measures the center points between tracking results and groundtruth, \emph{cannot} assess the accuracy regarding the target size and angle for the 9DoF 3D bounding boxes.

\vspace{0.45em}
\noindent
\textbf{3D SOT Tasks.} GSOT3D consists of sequences of multiple modalities, comprising \emph{point cloud}, \emph{RGB image}, and \emph{depth}. This allows research on various 3D tracking tasks, including the single-modal \emph{3D SOT on point cloud (PC)} \textbf{3D-SOT$_{\text{PC}}$}, and multi-modal \emph{3D SOT on RGB-PC} (\textbf{3D-SOT$_{\text{RGB-PC}}$}) and \emph{3D SOT on RGB-D} (\textbf{3D-SOT$_{\text{RGB-D}}$}). 

\begin{figure}[!t]
\centering
\includegraphics[width=0.98\linewidth]{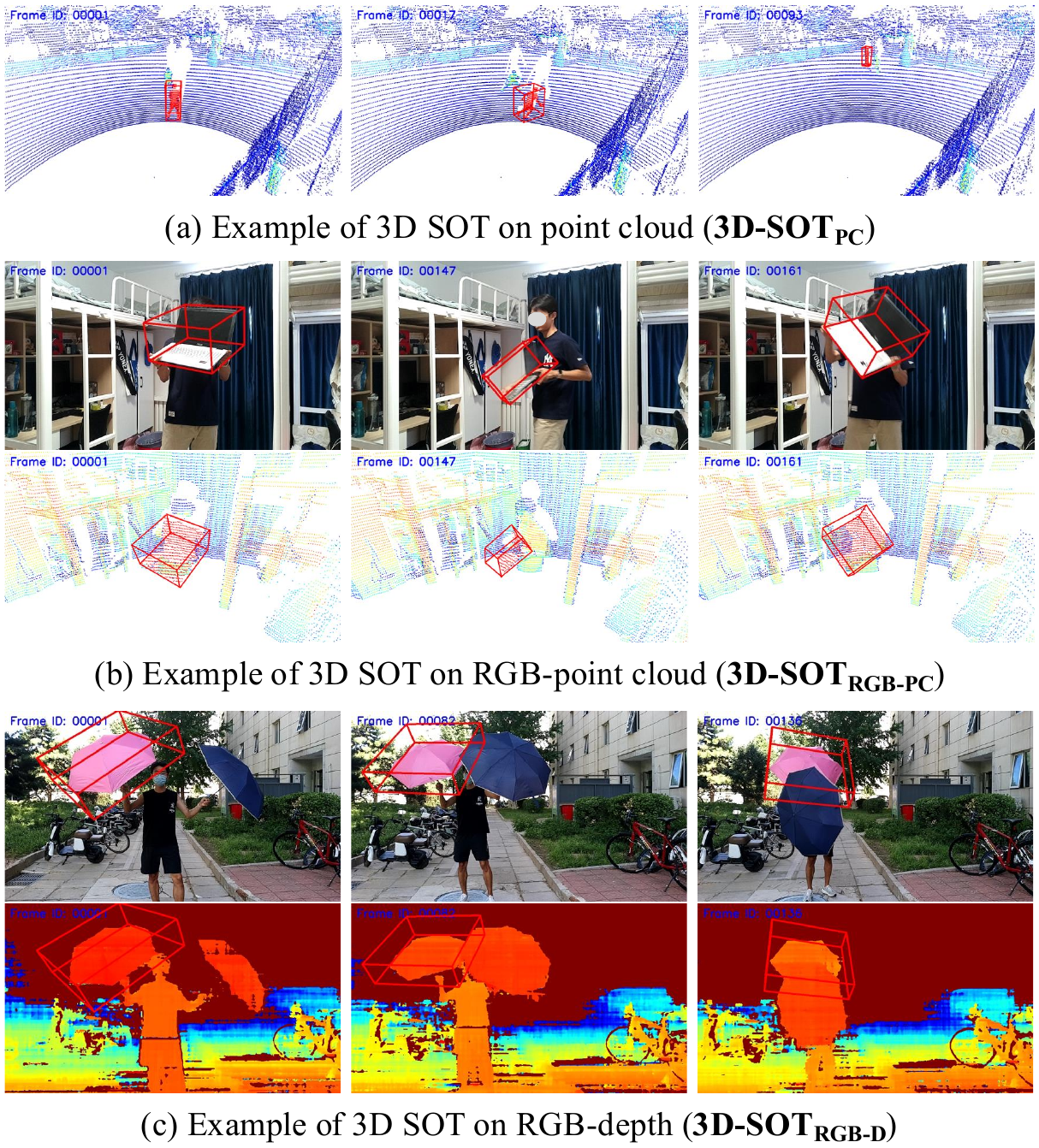}
\caption{Illustration of different 3D SOT tasks on GOST3D.}
\label{fig:task}
\vspace{-3.0mm}
\end{figure}

Given the initial 3D target box, 
3D-SOT$_{\text{PC}}$ aims to locate the target on the point clouds (see Fig.~\ref{fig:task} (a)), 3D-SOT$_{\text{RGB-PC}}$ localizes target object with point clouds and RGB images (see Fig.~\ref{fig:task} (b)), aiming to enhance the 3D tracking through appearance, and 3D-SOT$_{\text{RGB-D}}$ focuses on localizing the target using RGB and depth images (see Fig.~\ref{fig:task} (c)), providing a more cost-effective solution for 3D tracking. Due to limited space, please refer to our \textbf{supplementary material} for the detailed formulation of these tasks. 

For all tasks, except for used modalities, the dataset split and evaluation metric are the same. Please \textbf{\emph{note}}, since there are \emph{\textbf{very few}} trackers for 3D-SOT$_{\text{RGB-PC}}$ and 3D-SOT$_{\text{RGB-D}}$, we primarily focus on 3D-SOT$_{\text{PC}}$ in later baseline design and experiments due to more available trackers, and leave the study on SOT$_{\text{RGB-PC}}$ and 3D-SOT$_{\text{RGB-D}}$ to future work.

\section{The Proposed PROT3D}

\begin{figure}[!t]
    \centering
    \includegraphics[width=0.95\linewidth]{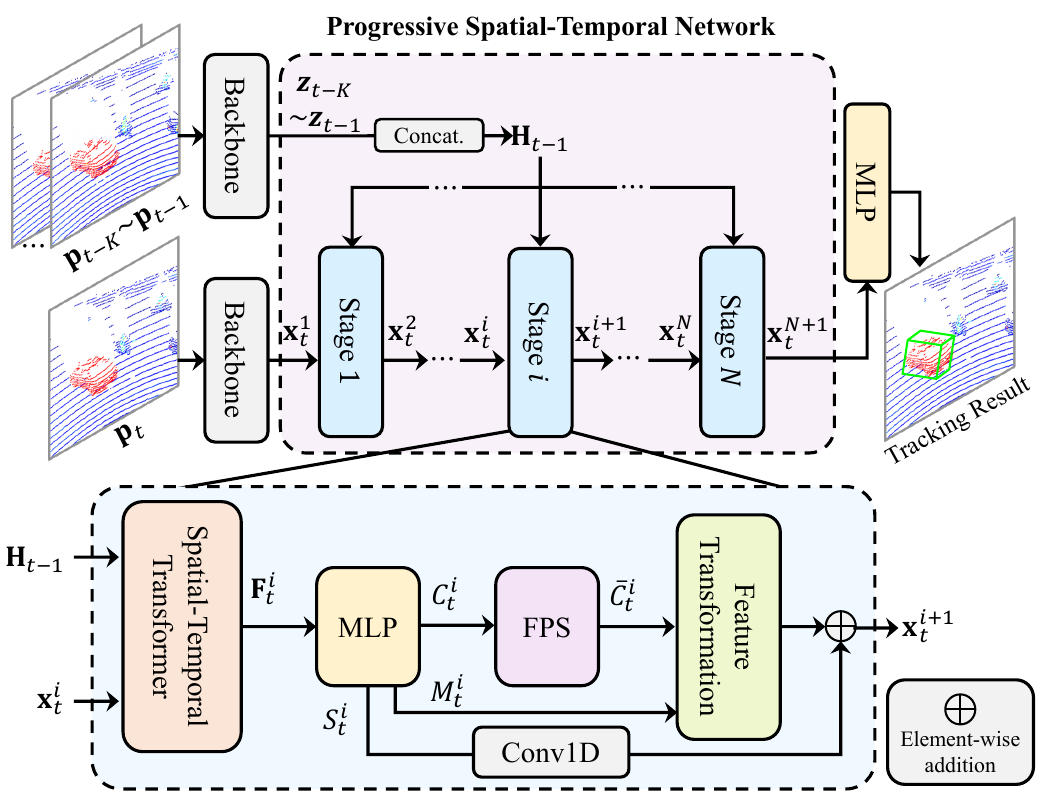}
    \caption{Architecture of the proposed PROT3D.}
    \label{fig:baseline}
    \vspace{-5.0mm}
\end{figure}

We present a simple yet effective tracker, PROT3D, for 3D-SOT$_{\text{PC}}$, as there are more available trackers for SOT$_{\text{PC}}$, and we will explore 3D-SOT$_{\text{RGB-PC}}$ and 3D-SOT$_{\text{RGB-D}}$ in the future. The key is to \emph{progressively} refine search region feature with multiple cascaded stages, as in Fig.~\ref{fig:baseline}. Each stage performs spatial-temporal target localization, and the result is used to augment the search region feature in the next stage.

Similar to~\cite{qi2020p2b}, PROT3D treats 3D tracking as a matching problem. Inspired by~\cite{xu2023mbptrack}, we leverage target cues from historical frames for robust performance. More specifically, given point cloud $\textbf{p}_{t}$ at frame $t$, we apply information from previous $K$ frames $\{\textbf{p}_{j}\}_{j=t-K}^{t-1}$ for tracking. We first extract their features through a shared backbone $\Phi(\cdot)$ as follows,
\begin{equation}
\setlength{\abovedisplayskip}{4pt} 
\setlength{\belowdisplayskip}{4pt}
\textbf{x}_{t}^{1}=\Phi(\textbf{p}_{t}) \;\;\;\;\;
    \textbf{z}_{j}=\Phi(\textbf{p}_{j}) \;\; j=t-K,\cdots, t-1
\end{equation}
where $\textbf{x}_{t}^{1}$ represents the feature of $\textbf{p}_t$ and $\textbf{z}_{j}$ is the feature of $\textbf{p}_{j}$ ($j=t-K,\cdots,t-1$). Then, we concatenate all features from historical frames via $\textbf{H}_{t-1}=\text{concat}(\textbf{z}_{t-K}, \cdots, \textbf{z}_{t-1})$ to obtain memory feature $\textbf{H}_{t-1}$ for frame $t$. After that, $\textbf{H}_{t-1}$ and $\textbf{x}_{t}^{1}$ are sent to the progressive spatial-temporal network with multiple stages, with each performing localization.

\begin{figure}[!t]
    \centering
    \includegraphics[width=0.95\linewidth]{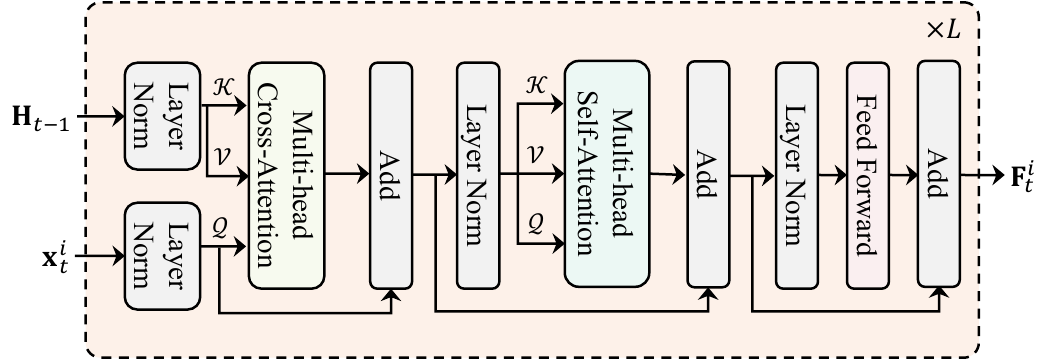}
    \caption{Architecture of spatial-temporal Transformer.}
    \label{fig:spt}
    \vspace{-4.0mm}
\end{figure}

Specifically, for stage $i$, it receives $\textbf{H}_{t-1}$ and $\textbf{x}_{t}^{i}$ as inputs. Then, a spatial-temporal Transformer is utilized to fuse the memory $\textbf{H}_{t-1}$ into $\textbf{x}_{t}^{i}$, as follows
\begin{equation}
\setlength{\abovedisplayskip}{4pt} 
\setlength{\belowdisplayskip}{4pt}
    \textbf{F}_{t}^{i} = \text{SPT}(\textbf{x}_{t}^{i}, \textbf{H}_{t-1})
\end{equation}
where $\textbf{F}_{t}^{i}$ is the feature after fusion. $\text{SPT}(\cdot,\cdot)$ represents the spatial-temporal Transformer, and comprises $L$ ($L$ is set to 2) layers. Similar to~\cite{xu2023mbptrack}, each layer consists of cross- and self-attention operations~\cite{vaswani2017attention} and a feed-forward network, as displayed in Fig.~\ref{fig:spt}. After that, $\textbf{F}_{t}^{i}$ is forwarded to a multi-layer perceptron (MLP) for localization, as follows
\begin{equation}
\setlength{\abovedisplayskip}{4pt} 
\setlength{\belowdisplayskip}{4pt}
    R_{t}^{i} = \text{MLP}(\textbf{F}_t^i)
\end{equation}
where $R_{t}^{i}=[C_t^i, M_t^i, S_t^i]$ is the localization result, with $C_t^i$ potential target center, $M_t^i$ targetness mask, and $S_t^i$ proposal scores. Then, we perform Farthest Point Sampling (FPS) on $C_t^i$ to refine point clouds, as follows
\begin{equation}
\setlength{\abovedisplayskip}{4pt} 
\setlength{\belowdisplayskip}{4pt} 
    \bar{C}_t^i = \text{FPS}(C_t^i)
\end{equation}
where $\bar{C}_t^i$ is sampled points. After FPS, the $\bar{C}_t^i$ and $M_t^i$ are fed to a feature transformation block (FTB) and the resulted feature is combined with the score information to generate the refined search region feature $\textbf{x}_{t}^{i+1}$, mathematically described as follows,
\begin{equation}\label{eqxt}
\setlength{\abovedisplayskip}{4pt} 
\setlength{\belowdisplayskip}{4pt}
    \textbf{x}_{t}^{i+1} = \text{FTB}(\bar{C}_t^i, M_t^i) + \text{Conv1D}(S_t^i)
\end{equation}
where $\text{FTB}(\cdot,\cdot)$ is feature transformation block, borrowed from~\cite{xu2023mbptrack}, and contains point-to-reference and a 3D convolution operation (see \textbf{supplementary material} for details). $\text{Conv1D}(\cdot)$ is 1D convolution to embed $S_t^i$ to score feature. 

Please note, $\textbf{x}_{t}^{i+1}$ in Eq.~(\ref{eqxt}) is generated by encoding target information $C_t^i$, $M_t^i$, and $S_t^i$, obtained via localization, and thus more discriminative for distinguishing target from background. For further refinement, $\textbf{x}_{t}^{i+1}$ is fed to the next stage ($i+1$), forming a progressive cascade architecture. This way, the search region feature can be gradually refined with more target cues, benefiting the final localization.


After the last $N^{\text{th}}$ stage, the generated $\textbf{x}_{t}^{N+1}$ is employed for final 9DoF target localization via MLP, as follows,
\begin{equation}
\setlength{\abovedisplayskip}{4pt} 
\setlength{\belowdisplayskip}{4pt}
    \mathcal{R}_{t} = \text{MLP}(\textbf{x}_{t}^{N+1})
\end{equation}
where $\mathcal{R}_{t}=[\mathcal{B}_{t}, \mathcal{S}_{t}] \in \mathbb{R}^{D\times 10}$, with $\mathcal{B}_{t} \in \mathbb{R}^{D\times 9}$ the 9DoF box parameters, $\mathcal{S}_{t} \in \mathbb{R}^{D\times 1}$ the targetness scores and $D$ the number of points in $\textbf{x}_{t}^{N+1}$. Finally, the tracking result $b_t$ is determined as follows,
\begin{equation}
\setlength{\abovedisplayskip}{4pt} 
\setlength{\belowdisplayskip}{4pt}
    b_t = \mathcal{B}_t(h) \;\;\; \text{where} \;\; h = \operatorname*{arg\,max}_{d=1,\cdots,D} \mathcal{S}(d)
\end{equation}
where $b_t=(x_t^*,y_t^*,x_t^*,\alpha_t^*,\beta_t^*,\gamma_t^*,l_t^*,h_t^*,w_t^*)$, predicting the translation offset $(x_t^*,y_t^*,x_t^*)$ of the center point and angle offset $(\alpha_t^*,\beta_t^*,\gamma_t^*)$ and size offset $(l_t^*,h_t^*,w_t^*)$ of target box from frame ($t-1$) to frame $t$.

\newcolumntype{C}[1]{>{\centering\arraybackslash}p{#1}}
\begin{table*}[!t]\small
  \centering
  \renewcommand{\arraystretch}{0.95}
  \tabcolsep=2mm
  \caption{Overall performance of eight state-of-the-art trackers and our PROT3D on 3D-SOT$_{\text{PC}}$ using mAO, mSR$_{50}$, and mSR$_{75}$. The best three results are highlighted in \myred{red}, \myblue{blue}, and \mygreen{green} fonts, respectively. Our PROT3D achieves the best results on all three metrics.}\vspace{-2mm}
  \resizebox{0.96\linewidth}{!}{
    \begin{tabular}
{rrC{1.1cm}@{}C{1.1cm}@{}C{1.1cm}@{}C{1.45cm}@{}C{1.3cm}@{}C{1.5cm}@{}C{1.5cm}@{}C{1cm}>{\columncolor{gray!15}}C{1.2cm}}
    \Xhline{1.2pt}
          &       & \tabincell{c}{P2B\\\cite{qi2020p2b}}   & \tabincell{c}{BAT\\\cite{zheng2021box}}   & \tabincell{c}{PTT\\\cite{shan2021ptt}}   & \tabincell{c}{M2-Track\\\cite{zheng2022beyond}} & \tabincell{c}{CXTrack\\\cite{xu2023cxtrack}} & \tabincell{c}{MBPTrack\\\cite{xu2023mbptrack}} & \tabincell{c}{SeqTrack-\\3D~\cite{lin2024seqtrack3d}} & \tabincell{c}{M3SOT\\\cite{liu2024m3sot}} & \tabincell{c}{\textbf{PROT3D}\\(ours)} \\
          \hline\hline
    \multicolumn{1}{r}{\multirow{3}[0]{*}{\tabincell{c}{w/ training\\ on GSOT3D}}} & mAO (\%)  & 9.79  & 6.56  & 14.00 & \mygreen{20.26} & 14.29 & \myblue{20.54} & 8.61  & 17.40 & \myred{21.97} \\
          & mSR$_{50}$ (\%) & 8.59  & 3.54  & 10.42 & \mygreen{14.34} & 8.39  & \myblue{16.55} & 5.25  & 12.47 & \myred{19.76} \\
          & mSR$_{75}$ (\%) & 1.75  & 0.88  & 1.60  & \mygreen{1.88}  & 1.02  & \myblue{2.57}  & 1.11  & 1.74  & \myred{5.22} \\
          \arrayrulecolor{gray}\hline
          \multicolumn{1}{r}{\multirow{3}[0]{*}{\textcolor[rgb]{ .651,  .651,  .651}{\tabincell{c}{w/o training \\on GSOT3D}}}} & \textcolor[rgb]{ .651,  .651,  .651}{mAO (\%)} & \textcolor[rgb]{ .651,  .651,  .651}{2.81} & \textcolor[rgb]{ .651,  .651,  .651}{1.91} & \textcolor[rgb]{ .651,  .651,  .651}{2.36} & \textcolor[rgb]{ .651,  .651,  .651}{3.65} & \textcolor[rgb]{ .651,  .651,  .651}{2.42} & \textcolor[rgb]{ .651,  .651,  .651}{3.38} & \textcolor[rgb]{ .651,  .651,  .651}{1.54} & \textcolor[rgb]{ .651,  .651,  .651}{2.68} & \textcolor[rgb]{ .651,  .651,  .651}{-} \\
          & \textcolor[rgb]{ .651,  .651,  .651}{mSR$_{50}$ (\%)} & \textcolor[rgb]{ .651,  .651,  .651}{1.35} & \textcolor[rgb]{ .651,  .651,  .651}{1.24} & \textcolor[rgb]{ .651,  .651,  .651}{1.29} & \textcolor[rgb]{ .651,  .651,  .651}{1.32} & \textcolor[rgb]{ .651,  .651,  .651}{1.19} & \textcolor[rgb]{ .651,  .651,  .651}{1.81} & \textcolor[rgb]{ .651,  .651,  .651}{0.90} & \textcolor[rgb]{ .651,  .651,  .651}{1.36} & \textcolor[rgb]{ .651,  .651,  .651}{-} \\
          & \textcolor[rgb]{ .651,  .651,  .651}{mSR$_{75}$ (\%)} & \textcolor[rgb]{ .651,  .651,  .651}{0.60} & \textcolor[rgb]{ .651,  .651,  .651}{0.60} & \textcolor[rgb]{ .651,  .651,  .651}{0.67} & \textcolor[rgb]{ .651,  .651,  .651}{0.61} & \textcolor[rgb]{ .651,  .651,  .651}{0.63} & \textcolor[rgb]{ .651,  .651,  .651}{0.65} & \textcolor[rgb]{ .651,  .651,  .651}{0.61} & \textcolor[rgb]{ .651,  .651,  .651}{0.62} & \textcolor[rgb]{ .651,  .651,  .651}{-} \\
          \Xhline{1.2pt}
    \end{tabular}%
    }
  \label{tab:overall}%
  \vspace{-3mm}
\end{table*}%

\begin{figure*}[!t]
    \centering
    \includegraphics[width=0.96\linewidth]{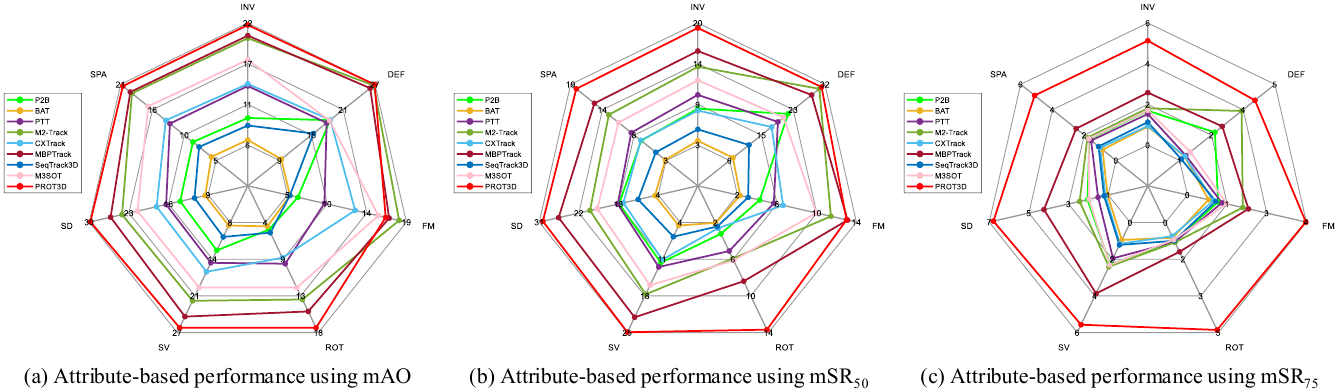}
    \caption{Attribute-based performance and comparison using mAO (image (a)), mSR$_{50}$ (image (b)), and mSR$_{75}$.}
    \label{fig:att}
    \vspace{-4.0mm}
\end{figure*}

Please note, PROT3D is a \emph{class-agnostic} 3D tracker that is able to track the target object of any categories. The loss of PROT3D is computed with loss function for final target estimation. Due to space limitation, please refer to our \textbf{supplementary material} for details of the loss function.

\vspace{0.3em}
\noindent
\textbf{Implementation.} PROT3D is implemented using PyTorch \cite{paszke2019pytorch}, and trained for 80 epochs using Adam~\cite{KingmaB14}. The initial learning rate is 0.001, and the batchsize is 9. In PROT3D, the number of stages is set to 2, and the memory size $K$ is set to 3. Our full code and model will be released.

\section{Experiments}

Please \textbf{note} again, we primary focus on experiments for 3D-SOT$_{\text{PC}}$ trackers, as most currently open-sourced 3D trackers with available implementations belong to 3D-SOT$_{\text{PC}}$. 

\vspace{0.3em}
\noindent
\textbf{Evaluated Trackers.} We evaluate eight representative 3D trackers that share their executable codes on GSOT3D, and provide basis for the future comparison, including P2B~\cite{qi2020p2b}, BAT~\cite{zheng2021box}, PTT~\cite{shan2021ptt}, M2-Track~\cite{zheng2022beyond}, CXTrack~\cite{xu2023cxtrack}, MBPTrack~\cite{xu2023mbptrack}, SeqTrack3D~\cite{lin2024seqtrack3d}, and M3SOT~\cite{liu2024m3sot}. The summary of these trackers is in the \textbf{supplementary material}. 


\subsection{Evaluation Results}

\textbf{Overall Performance.} We evaluate eight representative 3D trackers on 3D-SOT$_{\text{PC}}$ and the proposed PROT3D on test set of GSOT3D. Tab.~\ref{tab:overall} displays the results and comparison using mAO, mSR$_{50}$, and mSR$_{75}$. For the fair comparison, we retrain all evaluated trackers using training set of GSOT3D and compare them with our PROT3D in the Tab.~\ref{tab:overall}. We can observe that, PROT3D achieves the best result with 21.97\% mAO, 19.76\% mSR$_{50}$, and 5.22\% mSR$_{75}$, outperforming the second best MBPTrack with 20.54\% mAO by 1.43\%, 16.55\% mSR$_{50}$ by 3.21\%, and 2.57\% mSR$_{75}$ by 2.65\% and the third best M2-Track with 20.26\% mAO by 1.71\%, 14.34\% mSR$_{50}$ by 5.42, and 1.88\% mSR$_{75}$ by 3.34\%. This evidences the superiority of PROT3D with progressive refinement for more robust generic tracking. It is worth noting that, for all trackers, the mSR$_{75}$ score is much lower than the mSR$_{50}$ score, as mSR$_{75}$ has a higher threshold (0.75) than mSR$_{50}$ (0.5) and thus is more restrict.

Besides, Tab.~\ref{tab:overall} shows comparison of evaluated trackers using GSOT3D$_{\text{Tra}}$ or not for retraining. For the tracker that does not use GSOT3D$_{\text{Tra}}$ for training, we directly utilize its default model pre-trained from KITTI for evaluation. As in Tab.~\ref{tab:overall}, we observe that, re-training these trackers on GSOT3D can significantly improve their results on all three metrics. This shows the necessity of a more diverse dataset such as our GSOT3D for generic 3D object tracking.

\vspace{0.2em}
\noindent
\textbf{Attribute-based Performance.} In order to further analyze different algorithms, we conduct evaluation and comparison under seven attributes using mAO, mSR$_{50}$, and mSR$_{75}$. For fair comparison, all the compared trackers are trained using GSOT3D$_{\text{Tra}}$. Fig.~\ref{fig:att} reports the results. From Fig~\ref{fig:att}, we can see that, the proposed PROT3D achieves the best results on six out of seven attributes using mAO and mSR$_{50}$, and the best results on all seven attributes on all seven attributes using harder mSR$_{75}$. All these results show that, PROT3D is more robust and precise than other trackers in tracking.

Because of limited space, we demonstrate more qualitative results and analysis in the \textbf{supplementary material}.

\subsection{Comparison with Other Benchmark}

\begin{table}[!t]\small
  \centering
  \renewcommand{\arraystretch}{1.05}
  \tabcolsep=1.5mm
  \caption{Comparison of GSOT3D with KITTI.}\vspace{-2mm}
  \resizebox{0.99\linewidth}{!}{
    \begin{tabular}{rcccccc}
    \Xhline{1.2pt}
          & \multicolumn{3}{c}{\textbf{KITTI}~\cite{geiger2012we}} & \multicolumn{3}{c}{\textbf{GSOT3D} (ours)} \\
          \cmidrule(lr){2-4} \cmidrule(lr){5-7}
          & \tabincell{c}{mAO\\(\%)}   & \tabincell{c}{mSR$_{50}$\\(\%)} & \tabincell{c}{mSR$_{75}$ \\ (\%)} & \tabincell{c}{mAO\\(\%)}   & \tabincell{c}{mSR$_{50}$\\(\%)} & \tabincell{c}{mSR$_{75}$\\(\%)} \\
    \hline\hline
    P2B~\cite{qi2020p2b}   & 63.25 & 78.57 & 39.52 & 9.79  & 8.59  & 1.75 \\
    BAT~\cite{zheng2021box}   & 56.65 & 70.44 & 32.70  & 6.56  & 3.54  & 0.88 \\
    PTT~\cite{shan2021ptt}   & 52.30  & 66.32 & 40.79 & 14.00    & 10.42 & 1.60 \\
    M2-Track~\cite{zheng2022beyond} & 67.71 & 86.43 & 44.00    & 20.26 & 14.34 & 1.88 \\
    CXTrack~\cite{xu2023cxtrack} & 70.18 & 87.95 & 46.06 & 14.29 & 8.39  & 1.02 \\
    MBPTrack~\cite{xu2023mbptrack} & 71.95 & 90.50  & 51.54 & 20.54 & 16.55 & 2.57 \\
    SeqTrack3D~\cite{lin2024seqtrack3d} & 32.01 & 32.28 & 11.36 & 8.61  & 5.25  & 1.11 \\
    M3SOT~\cite{liu2024m3sot} & 64.58 & 81.33 & 35.38 & 17.40  & 12.47 & 1.74 \\
    \Xhline{1.2pt}
    \end{tabular}}
  \label{tab:dataset_comp}%
  \vspace{-3mm}
\end{table}%

KITTI~\cite{geiger2012we} is currently the most popular dataset for 3D SOT on point clouds. Nevertheless, as mentioned before, the sequences of KITTI are limited to only a few object categories and constrained traffic scenarios, making it not suitable for generic 3D object tracking. Compared to KITTI, GSOT3D includes more target classes from diverse environments. As a consequence, our GSOT3D is more challenging but realistic for real-world applications.

We conduct a comparison of our GSOT3D with KITTI. Tab.~\ref{tab:dataset_comp} reports the results of evaluated trackers on GSOT3D and KITTI using mAO, mSR$_{50}$, and mSR$_{75}$. As shown in Tab.~\ref{tab:dataset_comp}, we clearly see that, all current trackers suffer from a significant performance drop on GSOT3D, which 
 shows the challenges from more categories and diverse scenarios and more efforts are needed for generic 3D object tracking.

\subsection{Ablation Study on PROT3D}

\textbf{9DoF box prediction and progressive architecture.} Different from previous 3D trackers predicting a 7DoF bounding box, our PROT3D estimates a more precise 9DoF 3D bounding box as the tracking result. In addition, PROT3D applies a novel progressive architecture for tracking, which enables better features for robust localization. Tab.~\ref{tab:dof} lists the experiment results. The baseline (\ding{182}) contains one stage and predicts a 7DoF box, and achieves the mAO of 19.86\%, mSR$_{50}$ of 15.16\%, and mSR$_{75}$ of 2.36\%. When changing to the 9DoF box prediction (\ding{183}), the performance is improved to 20.03\% mAO, 15.46\% mSR$_{50}$, and 3.29\% mSR$_{75}$, showing effectiveness of using 9DoF for 3D tracking. It is worth noting, the gains by 9DoF are not very significant, as most objects in GSOT3D are rigid and only a small part of the sequences contain deformable objects. Nonetheless, in the real world, there exist more non-rigid objects, and 9DoF box prediction is still more desirable. When further applying our progressive architecture (\ding{184}), the results are largely boosted to 21.97\% mAO, 19.76\% mSR$_{50}$, 5.22\% mSR$_{75}$, which clearly validates the efficacy of our progressive refinement for generic 3D object tracking.  

\begin{table}[!t]\small
  \centering
  \renewcommand{\arraystretch}{0.95}
  \tabcolsep=1.8mm
  \caption{Analysis of 9DoF prediction and progressive architecture}\vspace{-2mm}
  \resizebox{0.75\linewidth}{!}{
    \begin{tabular}{cccccc}
    \Xhline{1.2pt}
          & \tabincell{c}{9DoF \\Box} & \tabincell{c}{Progressive \\Architecture}
          & \tabincell{c}{mAO\\(\%)}   & \tabincell{c}{mSR$_{50}$\\(\%)} & \tabincell{c}{mSR$_{75}$\\(\%)} \\
          \hline\hline
          \ding{182} &   -    &   -    & 19.86 & 15.16 & 2.36 \\
          \ding{183} &   \cmark   &   -    & 20.03 & 15.46 & 3.29 \\
          \ding{184} &   \cmark    &  \cmark     & \textbf{21.97} & \textbf{19.76} & \textbf{5.22} \\
    \Xhline{1.2pt}
    \end{tabular}
    }
  \label{tab:dof}%
\end{table}%

\begin{table}[!t]\small
  \centering
  \renewcommand{\arraystretch}{1}
  \tabcolsep=3mm
  \caption{Analysis of the number $N$ of stages in our PROT3D.}\vspace{-2mm}
  \resizebox{0.75\linewidth}{!}{
    \begin{tabular}{ccccc}
    \Xhline{1.2pt}
    &\tabincell{c}{Number of \\Stages} & \tabincell{c}{mAO\\(\%)}   & \tabincell{c}{mSR$_{50}$\\(\%)} & \tabincell{c}{mSR$_{75}$\\(\%)} \\
    \hline\hline
    \ding{182} &$N=1$     & 20.03 & 15.46 & 3.29 \\
    \ding{183} & $N=2$ & \textbf{21.97} & \textbf{19.76} & \textbf{5.22} \\
    \ding{184} & $N=3$     & 21.58 & 19.61 & 5.19 \\
    \Xhline{1.2pt}
    \end{tabular}%
    }
  \label{tab:stages}%
\end{table}%

\begin{table}[!t]\small
  \centering
  \renewcommand{\arraystretch}{1}
  \tabcolsep=3mm
  \caption{Analysis of the memory size $K$ in our PROT3D.}\vspace{-2mm}
  \resizebox{0.75\linewidth}{!}{
    \begin{tabular}{ccccc}
    \Xhline{1.2pt}
    & \tabincell{c}{Memory \\Size} & \tabincell{c}{mAO\\(\%)}   & \tabincell{c}{mSR$_{50}$\\(\%)} & \tabincell{c}{mSR$_{75}$\\(\%)} \\
    \hline\hline
    \ding{182}& $K=2$     & 21.37 & 19.52 & 5.32 \\
    \ding{183} & $K=3$ & \textbf{21.97} & \textbf{19.76} & \textbf{5.22} \\
    \ding{184} & $K=4$ & 21.84 & 19.69 & 5.17 \\
    \Xhline{1.2pt}
    \end{tabular}%
    }
  \label{tab:size}%
  \vspace{-3mm}
\end{table}%

\vspace{0.3em}
\noindent
\textbf{Number of progressive stages.} The core of our PROT3D is a progressive network with multiple stages of refinement. To explore the impact of number $N$ of stages in PROT3D, we conduct an ablation in Tab.~\ref{tab:stages}. We observe, when using two stages (\ding{183}), PROT3D shows the best results of 21.97\% mAO, 19.76 mSR$_{50}$, and 5.22\% mSR$_{75}$. When further increasing the number of stages to 3 (\ding{184}), the performance is slightly decreased. Thus, we set $N$ to 2 in this work.

\vspace{0.3em}
\noindent
\textbf{Memory size.} We adopt a memory containing previous $K$ frames for tracking. We ablate the memory size $K$ in Tab.~\ref{tab:size}. We observe that, when using 3 previous frames (\ding{183}) in the memory, PROT3D shows the best tracking performance.

\section{Conclusion and Limitation}

In this paper, we introduce GSOT3D, a new benchmark for generic 3D SOT. It contains 620 multimodal sequences with over 123K frames, and supports different 3D single object tracking tasks. To the best of our knowledge, GSOT3D is the largest benchmark to date dedicated to 3D SOT. Besides, we assess several representative trackers on GSOT3D to understand their performance and to offer comparison for future research. Furthermore, we present a simple yet effective progressive tracker PROT3D and obtain state-of-the-art result. We believe that, our benchmark, evaluation, and new baseline will inspire more research towards generic 3D object tracking and facilitate its real-world applications.

Despite contributions, there exist a few limitations. First, the experiments are mainly focused on the 3D-SOT$_{\text{PC}}$, and study on 3D-SOT$_{\text{RGB-PC}}$ and 3D-SOT$_{\text{RGB-D}}$ is not provided. Second, the sequences in GSOT3D are relatively short, and not suitable for long-term tracking. Given 3D-SOT$_{\text{PC}}$ is the current research focus and our major goal is to offer a new benchmark for generic tracking, we leave study of more 3D tracking tasks and long-term 3D tracking to the future work.

\noindent\section*{Supplementary Material}

\noindent
In this supplementary material, we present more details and analysis as well as results of our work, as follows,

\begin{itemize}
	\setlength{\itemsep}{2pt}
	\setlength{\parsep}{2pt}
	\setlength{\parskip}{2pt}
	
	\item[] \textbf{S1 \; Mobile Robotic Platform} \\ In this section, we demonstrate more details of our mobile robotic platform used for multimodal data collection.

	\item[] \textbf{S2 \; Annotation Tool} \\ We display more details of the annotation tool in labeling sequences with 9DoF 3D bounding boxes and its reliability analysis for high-quality annotation.
	

    \item[] \textbf{S3 \; More Statistics} \\ We demonstrate more statistics on GSOT3D regarding sequence length and per-category point density .

    \item[] \textbf{S4 \; Evaluation Metrics and 3D IoU} \\ We demonstrate detailed process on how to calculate the evaluation metrics and 3D IoU.
    
    \item[] \textbf{S5 \; Formulation of Different 3D SOT Tasks} \\ We describe the formulation of different 3D SOT tasks.

    \item[] \textbf{S6 \; Details of Feature Transformation Block} \\ We present the details of the feature transformation block adopted in our PROT3D.

    \item[] \textbf{S7 \; Loss Function} \\ We present details of the loss function to train PROT3D.
    
    \item[] \textbf{S8 \; Summary of Evaluated Trackers} \\ We offer a summary for  trackers assessed on GOST3D.

    \item[] \textbf{S9 \;Qualitative Results} \\ We offer more qualitative analysis of our PROT3D and its comparison to other trackers on GSOT3D.

    \item[] \textbf{S10 \; Maintenance and Responsible Usage of GSOT3D for Research} \\ We discuss the maintenance and responsible usage of our proposed GSOT3D for research.
	
\end{itemize}

\section*{S1 \; Mobile Robotic Platform}

\begin{figure}[!t]
    \centering
    \includegraphics[width=0.83\linewidth]{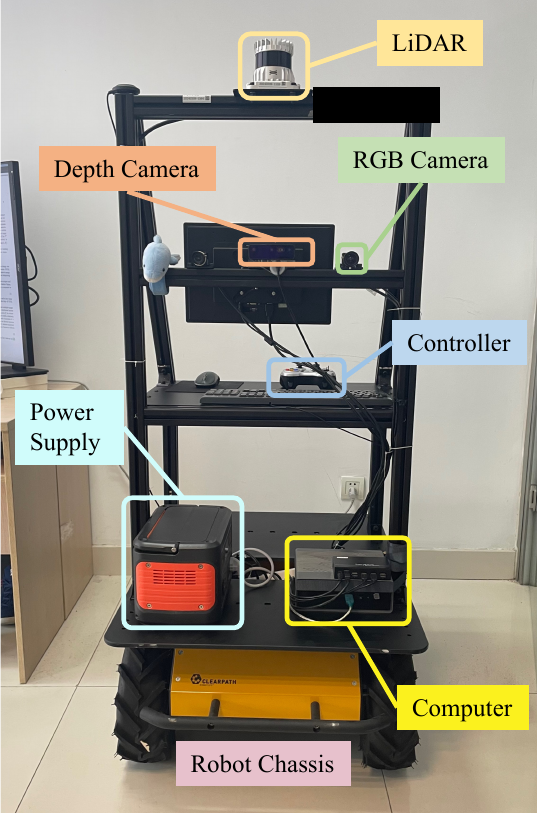}
    \caption{Our mobile robotic platform for data collection.}
    \label{fig:platform}
\end{figure}

To collect multimodal data for GSOT3D, we build a mobile robotic platform based on Clearpath Husky A200. Multiple sensors, including a 64-beam LiDAR, an RGB camera and a depth camera, are deployed on the platform with careful calibration using the tool from~\cite{dhall2017lidar}. Fig.~\ref{fig:platform} shows the picture of our mobile robotic platform for multimodal data acquisition in developing GSOT3D, and the specific configuration of sensors and robot chassis are listed in Tab.~\ref{tab:specitic_config}.

\begin{table}[!t]\small
    \centering
    \caption{Specific configuration of our mobile robotic platform.}
        \begin{tabular}{ll}
            \Xhline{1.2pt}
            \textbf{Device Name} & \textbf{Specification} \\ 
            \hline
            LiDAR Sensor       & Ouster OS-64 (64-beam)           \\
            Depth Camera       & OAK D-Pro              \\
            RGB Camera         & FLIR BFS-U3-32S4C-C    \\ 
            Robot   Chassis    & Clearpath Husky A200   \\
            \Xhline{1.2pt}
        \end{tabular}
    \label{tab:specitic_config}
\end{table}

\section*{S2 \; Annotation Tool}

\begin{figure*}[!t]
    \centering
    \includegraphics[width=0.88\linewidth]{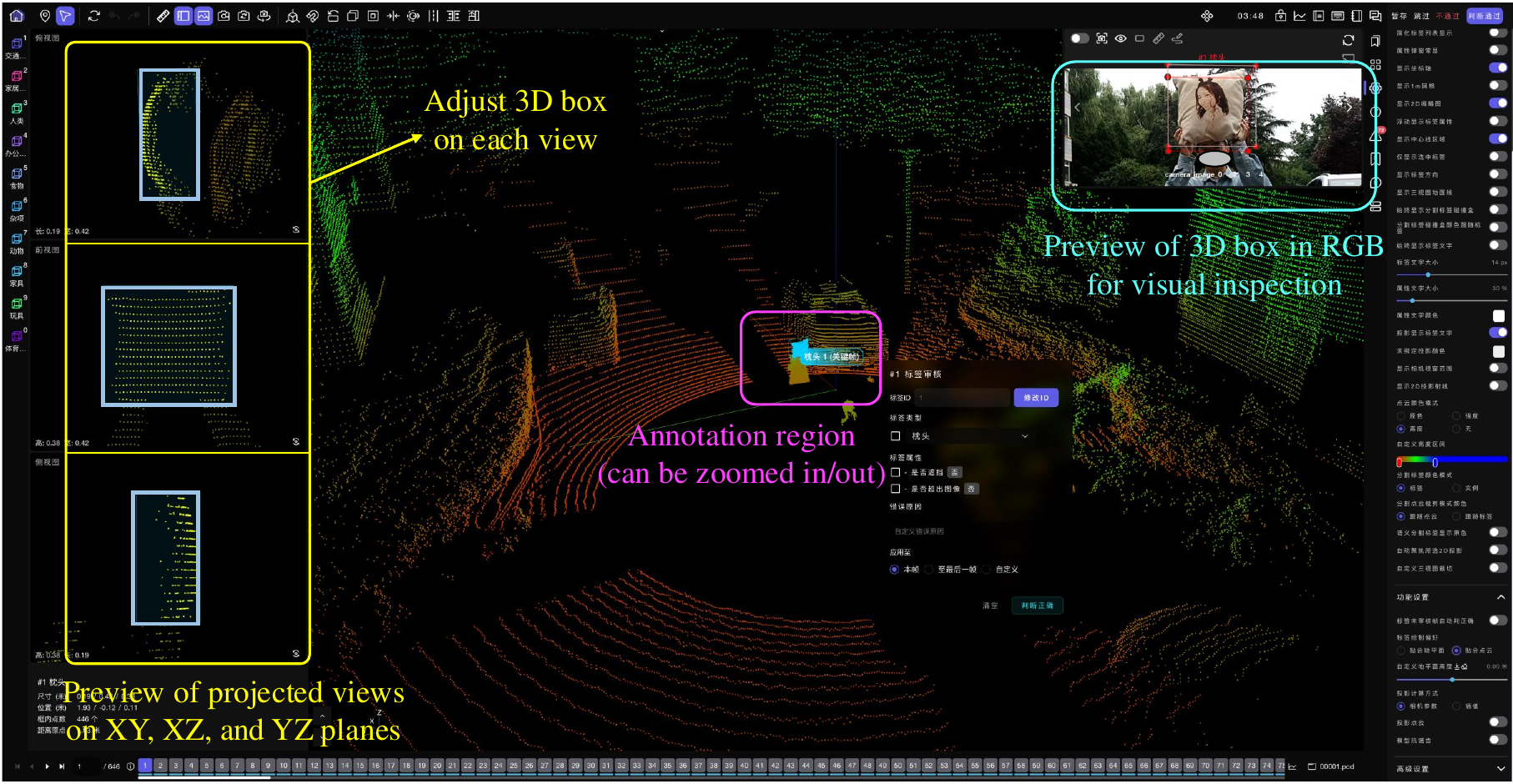}
    \caption{Annotation interface of our used annotation tool.}
    \label{fig:molardata}
\end{figure*}

\begin{figure*}[!t]
    \centering
    \includegraphics[width=0.88\linewidth]{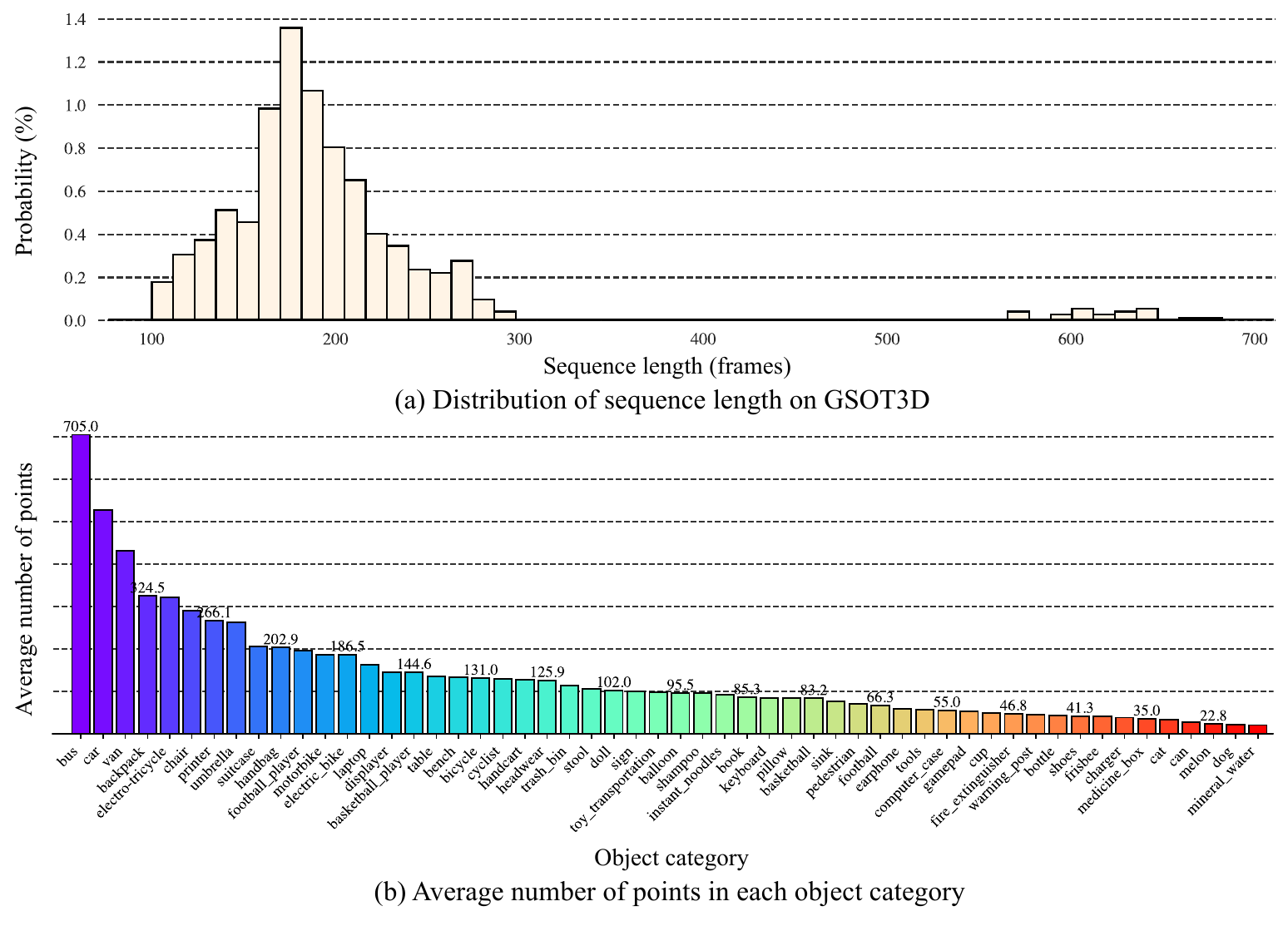}
    \vspace{-3mm}
    \caption{Statistics on GSOT3D. Image (a): Distribution of sequence length. Image (b): Average number of points in each object category}
    \label{fig:statistic}
\end{figure*}

For data labeling, we use the annotation tool provided by a company. Fig.~\ref{fig:molardata} shows the interface for 3D bounding box annotation. Specifically, for each point cloud frame, we perform initial annotation of the target object by drawing a 3D bounding box in the annotation region (note, this region can be flexibly zoomed in or out). Then, the initial 3D bounding box is refined by adjusting the 2D boxes on each projected view on XY, XZ, and YZ planes. In the annotation tool, a preview of the 3D box in the RGB image is provided for visual inspection of the refined box. By doing this, we can ensure the obtained annotation is reliable. Please note that, all the annotations from the labeler will be inspected careful by the experts (see this part in the main text) and further refined (by the same labeler) if necessary for high quality.

\section*{S3 \; More Statistics}

In this section, we demonstrate more statistics of GSOT3D. In specific, Fig.~\ref{fig:statistic} (a) shows distribution of sequence length on GSOT3D. Although the average length of GSOT3D is 198 frames, there exist several relatively longer ones with sequence length larger than 600 frames, which can be used for analyzing trackers on relatively longer sequences. Besides, Fig.~\ref{fig:statistic} (b) demonstrates the average number of points for each category. We can clearly see that, the categories of \emph{bus}, \emph{car}, and \emph{van} on average contain the most number of points, while the categories of \emph{dog} and \emph{mineral\_water} consist of the least number of points. We hope this statistics can help readers better understand our GSOT3D.

\section*{S4 \; Evaluation Metrics and 3D IoU}

Inspired by~\cite{huang2019got}, we utilize mean Average Overlap (\textbf{mAO}) and mean Success Rate (\textbf{mSR}) to measure different tracking algorithms. Specifically, mAO is calculated by averaging the class-wise overlaps, \ie, 3D  Intersection over Union (3D IoU, which will be detailed later), between all tracking results and the groundtruth, and mSR computes the class-wise percent of successful frames in which 3D IoU is larger than a threshold. mAO and mSR can be obtained as follows,
\begin{equation}
    \setlength{\abovedisplayskip}{4pt} 
    \setlength{\belowdisplayskip}{4pt}
    \begin{aligned}
    \text{mAO} & =\frac{1}{C} \sum_{c=1}^{C}\left(\frac{1}{\left|S_{c}\right|} \sum_{i \in S_{c}} \text{AO}_{i}\right) \\
    \text{mSR} & =\frac{1}{C} \sum_{c=1}^{C}\left(\frac{1}{\left|S_{c}\right|} \sum_{i \in S_{c}} \text{SR}_{i}\right)
    \end{aligned}
\end{equation}
where $C$ is the total number of object categories in GSOT3D, $S_c$ the set of all sequences belonging to category $c$. $\text{AO}_i$ represents the Average Overlap (AO) for the $i^{\text{th}}$ sequence in $S_c$, and $\text{SR}_i$ denotes Success Rate (SR). $\text{mSR}_{50}$ and $\text{mSR}_{75}$ refers to $\text{mSR}$ with thresholds of 0.5 and 0.75, respectively,  when computing success rate. 

\vspace{0.3em}
\noindent
\textbf{3D IoU.} Conventional 3D IoU often does not consider the targets that have symmetric structure. Nevertheless, in our GSOT3D, there exist many targets with symmetric structure, such as \emph{ball}, \emph{umbrella}, and so forth (148 sequences in total involved with symmetric structure). In these cases, conventional 3D IoU cannot be used for accurate measurement by considering a fixed direction. To deal with this, we leverage the strategy employed in~\cite{ahmadyan2021objectron,brazil2023omni3d} to calculate 3D IoU values between bounding boxes in arbitrary directions. Specifically, the predicted bounding box is rotated $k$ times along its axis of symmetry, and the prediction yielding the maximum 3D IoU among these $k$ rotations is selected as the final result. In our evaluation protocol, we set $k=120$, as this configuration achieves efficient computation while maintaining negligible error margins in the final measurement. The detailed calculation process can be seen in~\cite{ericson2004real}.

Therefore, for non-symmetric targets, we use method as in KITTI~\cite{geiger2012we} for 3D IoU calculation, while for symmetric targets, we use strategy as in~\cite{ahmadyan2021objectron,brazil2023omni3d} for 3D IoU computation.

\section*{S5 \; Formulation of Different 3D SOT Tasks}

GSOT3D is a unique platform to broaden research direction in 3D SOT by supporting different tasks, comprising single-modal 3D object tracking, \ie, \emph{3D SOT on Point Cloud (PC)} (3D-SOT$_{\text{PC}}$), and multi-modal 3D tracking, \ie, \emph{3D SOT on RGB-PC} (3D-SOT$_{\text{RGB-PC}}$) or \emph{RGB-Depth} (3D-SOT$_{\text{RGB-D}}$). 

\textbf{3D-SOT$_{\text{PC}}$} aims at locating the target object on the point clouds. Given the PC sequence and the initial 9DoF 3D target box, the goal is to estimate a set of 3D bounding boxes to represent the target positions in the sequence. This process can be formulated as follows,
\begin{equation}
    \setlength{\abovedisplayskip}{4pt} 
    \setlength{\belowdisplayskip}{4pt}
     \{b_i\}_{i=2}^{N} \leftarrow \mathcal{T}_{\text{PC}}(\{\textbf{p}_i\}_{i=1}^{N}, b_1)
\end{equation}
where $b_i=(x_i,y_i,z_i,w_i,h_i,l_i,\alpha_i,\beta_i,\gamma_i)$ is the 9DoF 3D box in frame $i$ $(1\le i \le N)$, with $(x_i,y_i,z_i)$, $(w_i,h_i,l_i)$, and $(\alpha_i,\beta_i,\gamma_i)$ the target position, scale, and rotation angle. $b_1$ is given in the first frame and $\{b_i\}_{i=2}^{N}$ are predicted by the tracker $\mathcal{T}_{\text{PC}}$. $\{\textbf{p}_i\}_{i=1}^{N}$ represent the PC sequence, and $N$ is the number of frames in the sequence.

Different from 3D-SOT$_{\text{PC}}$, \textbf{3D-SOT$_{\text{RGB-PC}}$} integrates the point clouds and RGB images for to locate target, aiming to improve 3D tracking using appearance information. It can be formulated as follows,
\begin{equation}\label{RGB-PC}
    \setlength{\abovedisplayskip}{4pt} 
    \setlength{\belowdisplayskip}{4pt}
     \{b_i\}_{i=2}^{N} \leftarrow \mathcal{T}_{\text{RGB-PC}}(\{\textbf{p}_i\}_{i=1}^{N}, \{I_i\}_{i=1}^{N}, b_1)
\end{equation}
where $b_1$ is the initial 9DoF 3D box, $\{b_i\}_{i=2}^{N}$ the predicted results by the tracker $\mathcal{T}_{\text{RGB-PC}}$, $\{\textbf{p}_i\}_{i=1}^{N}$ and $\{I_i\}_{i=1}^{N}$ the PC and RGB image sequences, respectively. 

Different than using PC, \textbf{3D-SOT$_{\text{RGB-D}}$} exploits a more economic way using RGB and depth images for 3D tracking, and can be formulated as follows,
\begin{equation}
    \setlength{\abovedisplayskip}{4pt} 
    \setlength{\belowdisplayskip}{4pt}
     \{b_i\}_{i=2}^{N} \leftarrow \mathcal{T}_{\text{RGB-D}}(\{D_i\}_{i=1}^{N}, \{I_i\}_{i=1}^{N}, b_1)
\end{equation}
where $\mathcal{T}_{\text{RGB-D}}$ denotes the 3D tracker, $\{D_i\}_{i=1}^{N}$ are the depth image sequence, and all others are the same as in Eq. (\ref{RGB-PC}).

By supporting different tracking tasks, GSOT3D expects to expand research directions in 3D SOT. 

\section*{S6 \; Details of Feature Transformation Block}

\begin{figure}[!t]
    \centering
    \includegraphics[width=0.95\linewidth]{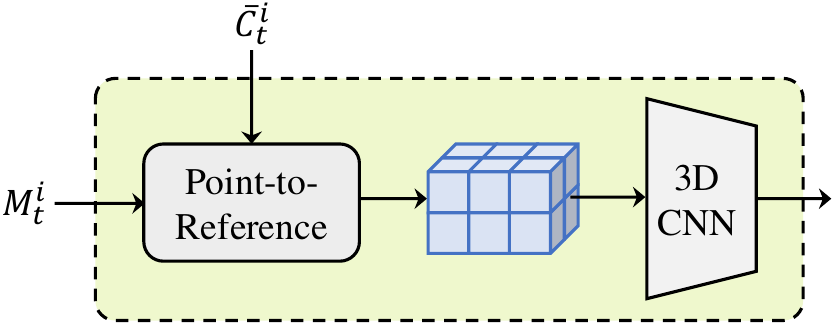}
    \caption{Architecture of the feature transformation block.}
    \label{fig:FTB}
\end{figure}

\begin{figure*}[!t]
    \centering
    \includegraphics[width=0.9\linewidth]{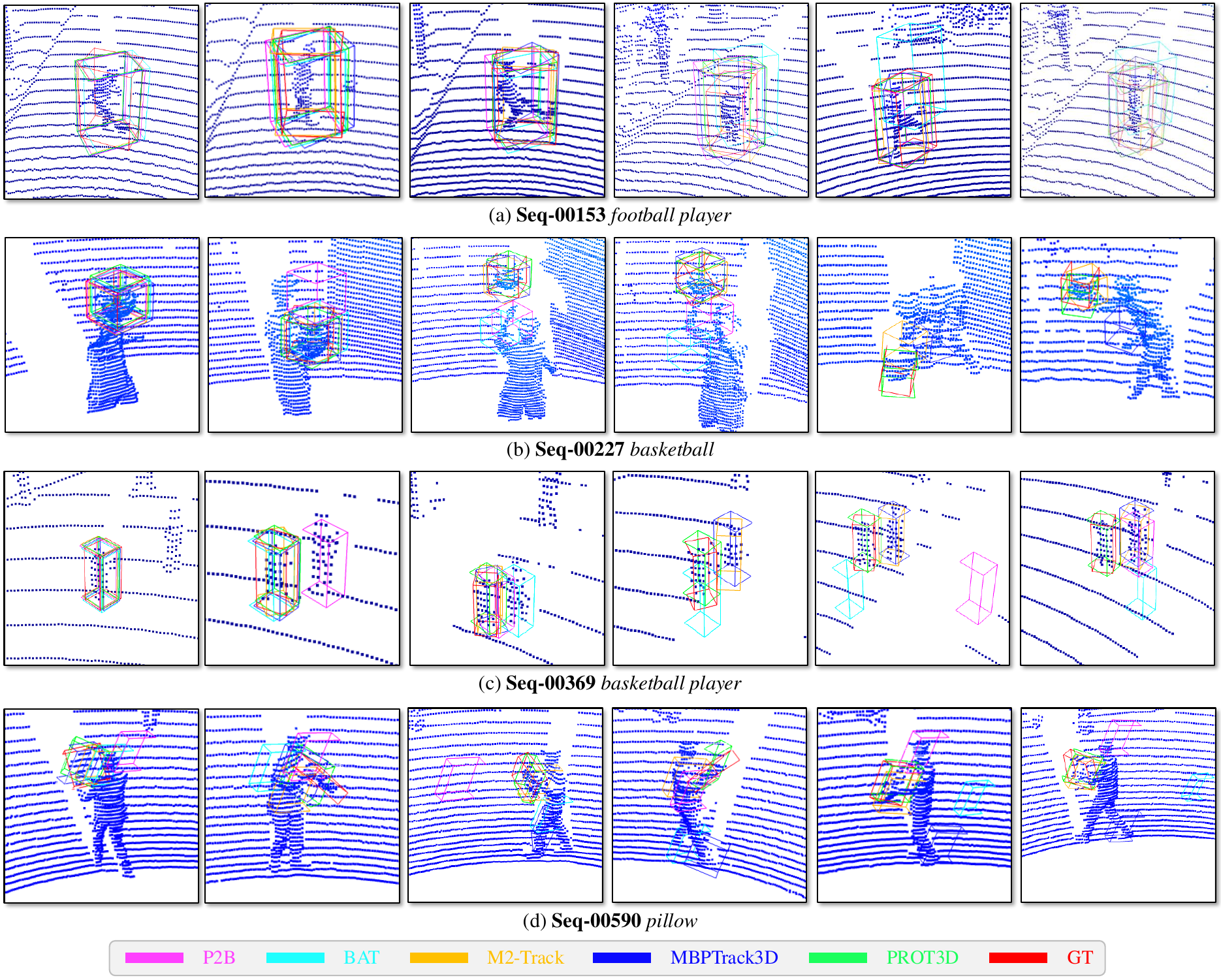}
    \caption{Qualitative results of several evaluated trackers and our proposed PROT3D. We can see that, the proposed PROT3D locates target object in different scenarios, showing its robustness for generic 3D object tracking.}
    \label{fig:pred_vis}
\end{figure*}

Fig.~\ref{fig:FTB} displays feature transformation block (FTB) applied in each stage of our PROT3D. The feature transformation block is borrowed from~\cite{xu2023mbptrack} for its effectiveness. In specific, we first send the targetness mask $M_t^i$ and the point feature $\bar{C}_{t}^{i}$ to the Point-to-Reference operation, which is composed of a concatenation operation, a MLP, and an EdgeConv layer~\cite{wang2019dynamic} for feature aggregation, as follows,
\begin{equation}
    \setlength{\abovedisplayskip}{4pt} 
    \setlength{\belowdisplayskip}{4pt}
    \begin{split}
        \hat{g}_t^{i}&=\text{Point-to-Reference}(\bar{C}_{t}^{i}, M_t^i) \\
        &=\text{EdgeConv}(\text{MLP}(\text{Concatenate}(\bar{C}_{t}^{i}, M_t^i)))
    \end{split}
\end{equation}
After this, the resulted feature $\hat{g}_t^{i}$ is fed into a 3D CNN network to generate point-wise feature. Fig.~\ref{fig:FTB} illustrates FTB. For more details, please kindly refer to~\cite{xu2023mbptrack}.

\section*{S7 \; Loss Function}

In this section, we present details regarding the loss function for training PROT3D. Specifically, after the $N^{\text{th}}$ stage, the final feature $\textbf{x}_{t}^{N+1}$ is sent to the MLP layer for prediction. Similar to previous work~\cite{xu2023mbptrack}, we use the following loss function for end-to-end training,
\begin{equation}
    \setlength{\abovedisplayskip}{4pt} 
    \setlength{\belowdisplayskip}{4pt}
    \mathcal{L}_\text{total}=\lambda_{\text{m}} \mathcal{L}_{\text{m}}+\lambda_{\text{c}} \mathcal{L}_{\text{c}}+\lambda_{\text{p}} \mathcal{L}_{\text{p}}+\lambda_{\text{s}} \mathcal{L}_{\text{s}}+\mathcal{L}_{\text{bbox}}
\end{equation}
where $\mathcal{L}_\text{total}$ represents the total training loss, $\mathcal{L}_\text{m}$ the standard cross-entropy loss to supervise the targetness mask, $\mathcal{L}_\text{c}$ the mean square loss to supervise the target center, $\mathcal{L}_\text{p}$ the cross-entropy loss to supervise proposal score, $\mathcal{L}_\text{s}$ the cross-entropy loss to supervise the targetness score $\mathcal{S}_\text{t}$, and $\mathcal{L}_{\text{bbox}}$ the smooth-L$_1$ loss to supervise the 9DoF box $\mathcal{B}_{t}$ (including 3D center offset and 6D pose offset of size and angle). $\lambda_{\text{m}}$, $\lambda_{\text{c}}$, $\lambda_{\text{p}}$, $\lambda_{\text{s}}$ are hyper-parameters to balance different losses and are set to 0.2, 10.0, 1.0, and 1.0, respectively.

Our code will be publicly released, and more details can be found in our implementation.

\begin{table}[!t]\small
  \centering
  \caption{Summary of evaluated trackers on GSOT3D.}
    \begin{tabular}{rccc}
    \Xhline{1.2pt}
    \textbf{Tracker} & \textbf{Where} & \textbf{Backbone} & \textbf{Transformer} \\
    \hline\hline
    P2B~\cite{qi2020p2b}                    & \multicolumn{1}{l}{CVPR'20} & PointNet++      & \xmark \\
    BAT~\cite{zheng2021box}                 & \multicolumn{1}{l}{ICCV'21} & PointNet++      & \xmark \\
    PTT~\cite{shan2021ptt}                  & \multicolumn{1}{l}{IROS'21} & PointNet++      & \cmark \\
    M2-Track~\cite{zheng2022beyond}         & \multicolumn{1}{l}{CVPR'22} & PointNet        & \xmark \\
    CXTrack~\cite{xu2023cxtrack}            & \multicolumn{1}{l}{CVPR'23} & DGCNN           & \cmark \\
    MBPTrack~\cite{xu2023mbptrack}          & \multicolumn{1}{l}{ICCV'23} & DGCNN           & \cmark \\
    SeqTrack3D~\cite{lin2024seqtrack3d}     & \multicolumn{1}{l}{ICRA'24} & PointNet++      & \cmark \\
    MS3SOT~\cite{liu2024m3sot}              & \multicolumn{1}{l}{AAAI'24} & DGCNN           & \cmark \\
    \Xhline{1.2pt}
    \end{tabular}%
  \label{tab:trackers}%
\end{table}%

\section*{S8 \; Summary of Evaluated Trackers}

To understand how existing trackers perform on GSOT3D and to provide comparison for future research, we assess eight representative trackers, including P2B~\cite{qi2020p2b}, BAT~\cite{zheng2021box}, PTT~\cite{shan2021ptt}, M2-Track~\cite{zheng2022beyond}, CXTrack~\cite{xu2023cxtrack}, MBPTrack~\cite{xu2023mbptrack}, SeqTrack3D~\cite{lin2024seqtrack3d}, and M3SOT~\cite{liu2024m3sot}. Please note that, these evaluated 3D trackers are point cloud-based, as almost all current 3D object trackers that share their implementations belong to this category. Tab.~\ref{tab:trackers} summarizes these trackers.

\section*{S9 \; Qualitative Results}

In this section, we show qualitative results of different trackers and our PROT3D on GSOT3D in Fig.~\ref{fig:pred_vis}. From Fig.~\ref{fig:pred_vis}, we can see that, existing state-of-the-art trackers such as M2-Track, MBPTrack fail to accurately localize the target object in challenging scenarios with frequent occlusions and similar distractors, while our PROT3D can robustly locate the target in these cases owing to its progressive refinement strategy, showing its efficacy for generic 3D tracking.

\section*{S10 \; Maintenance and Responsible Usage of GSOT3D for Research}

\textbf{Maintenance.} Our GSOT3D will be hosted on the popular Github (all download links and our models will be publicly released). This enables conveniently checking the feedback from the community, and thus allows for improvements via necessary maintenance and updates by the authors. Besides, the authors will try their best to collect evaluation results of future trackers, aiming at providing up-to-date analysis and comparison on GSOT3D. Our ultimate goal is to develop a long-term and stable platform for 3D object tracking.

\vspace{0.3em}
\noindent
\textbf{Responsible Usage of GSOT3D.} GSOT3D aims to facilitate research and applications of 3D single object tracking. It is developed and used for \textbf{\emph{research purpose only}}.

{
\small
\bibliographystyle{ieeenat_fullname}
\bibliography{main}

\begin{thebibliography}{41}
\providecommand{\natexlab}[1]{#1}
\providecommand{\url}[1]{\texttt{#1}}
\expandafter\ifx\csname urlstyle\endcsname\relax
  \providecommand{\doi}[1]{doi: #1}\else
  \providecommand{\doi}{doi: \begingroup \urlstyle{rm}\Url}\fi

\bibitem[Ahmadyan et~al.(2021)Ahmadyan, Zhang, Ablavatski, Wei, and Grundmann]{ahmadyan2021objectron}
Adel Ahmadyan, Liangkai Zhang, Artsiom Ablavatski, Jianing Wei, and Matthias Grundmann.
\newblock Objectron: A large scale dataset of object-centric videos in the wild with pose annotations.
\newblock In \emph{CVPR}, 2021.

\bibitem[Asvadi et~al.(2016)Asvadi, Girao, Peixoto, and Nunes]{asvadi20163d}
Alireza Asvadi, Pedro Girao, Paulo Peixoto, and Urbano Nunes.
\newblock 3d object tracking using rgb and lidar data.
\newblock In \emph{ITSC}, 2016.

\bibitem[Bibi et~al.(2016)Bibi, Zhang, and Ghanem]{bibi20163d}
Adel Bibi, Tianzhu Zhang, and Bernard Ghanem.
\newblock 3d part-based sparse tracker with automatic synchronization and registration.
\newblock In \emph{CVPR}, 2016.

\bibitem[Brazil et~al.(2023)Brazil, Kumar, Straub, Ravi, Johnson, and Gkioxari]{brazil2023omni3d}
Garrick Brazil, Abhinav Kumar, Julian Straub, Nikhila Ravi, Justin Johnson, and Georgia Gkioxari.
\newblock Omni3d: A large benchmark and model for 3d object detection in the wild.
\newblock In \emph{CVPR}, 2023.

\bibitem[Caesar et~al.(2020)Caesar, Bankiti, Lang, Vora, Liong, Xu, Krishnan, Pan, Baldan, and Beijbom]{caesar2020nuscenes}
Holger Caesar, Varun Bankiti, Alex~H Lang, Sourabh Vora, Venice~Erin Liong, Qiang Xu, Anush Krishnan, Yu Pan, Giancarlo Baldan, and Oscar Beijbom.
\newblock nuscenes: A multimodal dataset for autonomous driving.
\newblock In \emph{CVPR}, 2020.

\bibitem[Dhall et~al.(2017)Dhall, Chelani, Radhakrishnan, and Krishna]{dhall2017lidar}
Ankit Dhall, Kunal Chelani, Vishnu Radhakrishnan, and K~Madhava Krishna.
\newblock Lidar-camera calibration using 3d-3d point correspondences.
\newblock \emph{arXiv}, 2017.

\bibitem[Ericson(2004)]{ericson2004real}
Christer Ericson.
\newblock \emph{Real-time collision detection}.
\newblock Crc Press, 2004.

\bibitem[Fan et~al.(2019)Fan, Lin, Yang, Chu, Deng, Yu, Bai, Xu, Liao, and Ling]{fan2019lasot}
Heng Fan, Liting Lin, Fan Yang, Peng Chu, Ge Deng, Sijia Yu, Hexin Bai, Yong Xu, Chunyuan Liao, and Haibin Ling.
\newblock Lasot: A high-quality benchmark for large-scale single object tracking.
\newblock In \emph{CVPR}, 2019.

\bibitem[Fan et~al.(2021)Fan, Bai, Lin, Yang, Chu, Deng, Yu, Harshit, Huang, Liu, et~al.]{fan2021lasot}
Heng Fan, Hexin Bai, Liting Lin, Fan Yang, Peng Chu, Ge Deng, Sijia Yu, Harshit, Mingzhen Huang, Juehuan Liu, et~al.
\newblock Lasot: A high-quality large-scale single object tracking benchmark.
\newblock \emph{International Journal of Computer Vision}, 129:\penalty0 439--461, 2021.

\bibitem[Galoogahi et~al.(2017)Galoogahi, Fagg, Huang, Ramanan, and Lucey]{galoogahi2017need}
Hamed~Kiani Galoogahi, Ashton Fagg, Chen Huang, Deva Ramanan, and Simon Lucey.
\newblock Need for speed: A benchmark for higher frame rate object tracking.
\newblock In \emph{ICCV}, 2017.

\bibitem[Geiger et~al.(2012)Geiger, Lenz, and Urtasun]{geiger2012we}
Andreas Geiger, Philip Lenz, and Raquel Urtasun.
\newblock Are we ready for autonomous driving? the kitti vision benchmark suite.
\newblock In \emph{CVPR}, 2012.

\bibitem[Giancola et~al.(2019)Giancola, Zarzar, and Ghanem]{giancola2019leveraging}
Silvio Giancola, Jesus Zarzar, and Bernard Ghanem.
\newblock Leveraging shape completion for 3d siamese tracking.
\newblock In \emph{CVPR}, 2019.

\bibitem[Guo et~al.(2022)Guo, Mao, Zhou, Wang, and Li]{guo2022cmt}
Zhiyang Guo, Yunyao Mao, Wengang Zhou, Min Wang, and Houqiang Li.
\newblock Cmt: Context-matching-guided transformer for 3d tracking in point clouds.
\newblock In \emph{ECCV}, 2022.

\bibitem[Huang et~al.(2021)Huang, Zhao, and Huang]{huang2019got}
Lianghua Huang, Xin Zhao, and Kaiqi Huang.
\newblock Got-10k: A large high-diversity benchmark for generic object tracking in the wild.
\newblock \emph{IEEE Transactions on Pattern Analysis and Machine Intelligence}, 43\penalty0 (5):\penalty0 1562--1577, 2021.

\bibitem[Hui et~al.(2021)Hui, Wang, Cheng, Xie, and Yang]{hui20213d}
Le Hui, Lingpeng Wang, Mingmei Cheng, Jin Xie, and Jian Yang.
\newblock 3d siamese voxel-to-bev tracker for sparse point clouds.
\newblock In \emph{NeurIPS}, 2021.

\bibitem[Hui et~al.(2022)Hui, Wang, Tang, Lan, Xie, and Yang]{hui20223d}
Le Hui, Lingpeng Wang, Linghua Tang, Kaihao Lan, Jin Xie, and Jian Yang.
\newblock 3d siamese transformer network for single object tracking on point clouds.
\newblock In \emph{ECCV}, 2022.

\bibitem[Kingma and Ba(2015)]{KingmaB14}
Diederik~P. Kingma and Jimmy Ba.
\newblock Adam: {A} method for stochastic optimization.
\newblock In \emph{ICLR}, 2015.

\bibitem[Kristan et~al.(2016)Kristan, Matas, Leonardis, Voj{\'\i}{\v{r}}, Pflugfelder, Fernandez, Nebehay, Porikli, and {\v{C}}ehovin]{kristan2016novel}
Matej Kristan, Jiri Matas, Ale{\v{s}} Leonardis, Tom{\'a}{\v{s}} Voj{\'\i}{\v{r}}, Roman Pflugfelder, Gustavo Fernandez, Georg Nebehay, Fatih Porikli, and Luka {\v{C}}ehovin.
\newblock A novel performance evaluation methodology for single-target trackers.
\newblock \emph{IEEE Transactions on Pattern Analysis and Machine Intelligence}, 38\penalty0 (11):\penalty0 2137--2155, 2016.

\bibitem[Li et~al.(2015)Li, Lin, Wu, Yang, and Yan]{li2015nus}
Annan Li, Min Lin, Yi Wu, Ming-Hsuan Yang, and Shuicheng Yan.
\newblock Nus-pro: A new visual tracking challenge.
\newblock \emph{IEEE Transactions on Pattern Analysis and Machine Intelligence}, 38\penalty0 (2):\penalty0 335--349, 2015.

\bibitem[Liang et~al.(2015)Liang, Blasch, and Ling]{liang2015encoding}
Pengpeng Liang, Erik Blasch, and Haibin Ling.
\newblock Encoding color information for visual tracking: Algorithms and benchmark.
\newblock \emph{IEEE Transactions on Image Processing}, 24\penalty0 (12):\penalty0 5630--5644, 2015.

\bibitem[Lin et~al.(2024)Lin, Li, Cui, and Fang]{lin2024seqtrack3d}
Yu Lin, Zhiheng Li, Yubo Cui, and Zheng Fang.
\newblock Seqtrack3d: Exploring sequence information for robust 3d point cloud tracking.
\newblock In \emph{ICRA}, 2024.

\bibitem[Liu et~al.(2024)Liu, Wu, Gong, Miao, Ma, Xu, and Qin]{liu2024m3sot}
Jiaming Liu, Yue Wu, Maoguo Gong, Qiguang Miao, Wenping Ma, Cai Xu, and Can Qin.
\newblock M3sot: Multi-frame, multi-field, multi-space 3d single object tracking.
\newblock In \emph{AAAI}, 2024.

\bibitem[Ma et~al.(2023)Ma, Wang, Xiao, Wu, and Liu]{ma2023synchronize}
Teli Ma, Mengmeng Wang, Jimin Xiao, Huifeng Wu, and Yong Liu.
\newblock Synchronize feature extracting and matching: A single branch framework for 3d object tracking.
\newblock In \emph{ICCV}, 2023.

\bibitem[Muller et~al.(2018)Muller, Bibi, Giancola, Alsubaihi, and Ghanem]{muller2018trackingnet}
Matthias Muller, Adel Bibi, Silvio Giancola, Salman Alsubaihi, and Bernard Ghanem.
\newblock Trackingnet: A large-scale dataset and benchmark for object tracking in the wild.
\newblock In \emph{ECCV}, 2018.

\bibitem[Nie et~al.(2024)Nie, He, Lv, Zhou, Chae, and Xie]{nie2024towards}
Jiahao Nie, Zhiwei He, Xudong Lv, Xueyi Zhou, Dong-Kyu Chae, and Fei Xie.
\newblock Towards category unification of 3d single object tracking on point clouds.
\newblock In \emph{ICLR}, 2024.

\bibitem[Paszke et~al.(2019)Paszke, Gross, Massa, Lerer, Bradbury, Chanan, Killeen, Lin, Gimelshein, Antiga, et~al.]{paszke2019pytorch}
Adam Paszke, Sam Gross, Francisco Massa, Adam Lerer, James Bradbury, Gregory Chanan, Trevor Killeen, Zeming Lin, Natalia Gimelshein, Luca Antiga, et~al.
\newblock Pytorch: An imperative style, high-performance deep learning library.
\newblock In \emph{NeurIPS}, 2019.

\bibitem[Peng et~al.(2024)Peng, Gao, Liu, Li, Dong, Zhang, Fan, and Zhang]{peng2024vasttrack}
Liang Peng, Junyuan Gao, Xinran Liu, Weihong Li, Shaohua Dong, Zhipeng Zhang, Heng Fan, and Libo Zhang.
\newblock Vasttrack: Vast category visual object tracking.
\newblock In \emph{NeurIPS}, 2024.

\bibitem[Qi et~al.(2020)Qi, Feng, Cao, Zhao, and Xiao]{qi2020p2b}
Haozhe Qi, Chen Feng, Zhiguo Cao, Feng Zhao, and Yang Xiao.
\newblock P2b: Point-to-box network for 3d object tracking in point clouds.
\newblock In \emph{CVPR}, 2020.

\bibitem[Shan et~al.(2021)Shan, Zhou, Fang, and Cui]{shan2021ptt}
Jiayao Shan, Sifan Zhou, Zheng Fang, and Yubo Cui.
\newblock Ptt: Point-track-transformer module for 3d single object tracking in point clouds.
\newblock In \emph{IROS}, 2021.

\bibitem[Vaswani et~al.(2017)Vaswani, Shazeer, Parmar, Uszkoreit, Jones, Gomez, Kaiser, and Polosukhin]{vaswani2017attention}
Ashish Vaswani, Noam Shazeer, Niki Parmar, Jakob Uszkoreit, Llion Jones, Aidan~N Gomez, {\L}ukasz Kaiser, and Illia Polosukhin.
\newblock Attention is all you need.
\newblock In \emph{NIPS}, 2017.

\bibitem[Wang et~al.(2021)Wang, Shu, Zhang, Jiang, Wang, Tian, and Wu]{wang2021towards}
Xiao Wang, Xiujun Shu, Zhipeng Zhang, Bo Jiang, Yaowei Wang, Yonghong Tian, and Feng Wu.
\newblock Towards more flexible and accurate object tracking with natural language: Algorithms and benchmark.
\newblock In \emph{CVPR}, 2021.

\bibitem[Wang et~al.(2019)Wang, Sun, Liu, Sarma, Bronstein, and Solomon]{wang2019dynamic}
Yue Wang, Yongbin Sun, Ziwei Liu, Sanjay~E Sarma, Michael~M Bronstein, and Justin~M Solomon.
\newblock Dynamic graph cnn for learning on point clouds.
\newblock \emph{ACM Transactions on Graphics}, 38\penalty0 (5):\penalty0 1--12, 2019.

\bibitem[Wu et~al.(2023)Wu, Xia, Wan, and Chan]{wuboosting}
Qiangqiang Wu, Yan Xia, Jia Wan, and Antoni~B Chan.
\newblock Boosting 3d single object tracking with 2d matching distillation and 3d pre-training.
\newblock In \emph{ECCV}, 2023.

\bibitem[Wu et~al.(2024)Wu, Sun, An, Salzmann, Zhang, and Yang]{wu20253d}
Qiao Wu, Kun Sun, Pei An, Mathieu Salzmann, Yanning Zhang, and Jiaqi Yang.
\newblock 3d single-object tracking in point clouds with high temporal variation.
\newblock In \emph{ECCV}, 2024.

\bibitem[Wu et~al.(2013)Wu, Lim, and Yang]{wu2013online}
Yi Wu, Jongwoo Lim, and Ming-Hsuan Yang.
\newblock Online object tracking: A benchmark.
\newblock In \emph{CVPR}, 2013.

\bibitem[Xu et~al.(2023{\natexlab{a}})Xu, Guo, Lai, and Zhang]{xu2023cxtrack}
Tian-Xing Xu, Yuan-Chen Guo, Yu-Kun Lai, and Song-Hai Zhang.
\newblock Cxtrack: Improving 3d point cloud tracking with contextual information.
\newblock In \emph{CVPR}, 2023{\natexlab{a}}.

\bibitem[Xu et~al.(2023{\natexlab{b}})Xu, Guo, Lai, and Zhang]{xu2023mbptrack}
Tian-Xing Xu, Yuan-Chen Guo, Yu-Kun Lai, and Song-Hai Zhang.
\newblock Mbptrack: Improving 3d point cloud tracking with memory networks and box priors.
\newblock In \emph{ICCV}, 2023{\natexlab{b}}.

\bibitem[Yang et~al.(2022)Yang, Zhang, Li, Chang, Leonardis, and Zheng]{yang2022towards}
Jinyu Yang, Zhongqun Zhang, Zhe Li, Hyung~Jin Chang, Ale{\v{s}} Leonardis, and Feng Zheng.
\newblock Towards generic 3d tracking in rgbd videos: Benchmark and baseline.
\newblock In \emph{ECCV}, 2022.

\bibitem[Zheng et~al.(2021)Zheng, Yan, Gao, Zhao, Zhang, Li, and Cui]{zheng2021box}
Chaoda Zheng, Xu Yan, Jiantao Gao, Weibing Zhao, Wei Zhang, Zhen Li, and Shuguang Cui.
\newblock Box-aware feature enhancement for single object tracking on point clouds.
\newblock In \emph{ICCV}, 2021.

\bibitem[Zheng et~al.(2022)Zheng, Yan, Zhang, Wang, Cheng, Cui, and Li]{zheng2022beyond}
Chaoda Zheng, Xu Yan, Haiming Zhang, Baoyuan Wang, Shenghui Cheng, Shuguang Cui, and Zhen Li.
\newblock Beyond 3d siamese tracking: A motion-centric paradigm for 3d single object tracking in point clouds.
\newblock In \emph{CVPR}, 2022.

\bibitem[Zhou et~al.(2022)Zhou, Luo, Luo, Liu, Pan, Cai, Zhao, and Lu]{zhou2022pttr}
Changqing Zhou, Zhipeng Luo, Yueru Luo, Tianrui Liu, Liang Pan, Zhongang Cai, Haiyu Zhao, and Shijian Lu.
\newblock Pttr: Relational 3d point cloud object tracking with transformer.
\newblock In \emph{CVPR}, 2022.

\end{thebibliography}
}


\end{document}